\documentclass[11pt,a4paper]{article}
\usepackage[T1]{fontenc}
\usepackage[utf8]{inputenc}
\usepackage{mathptmx}
\usepackage[scaled=.95]{helvet}
\usepackage{courier}
\usepackage[margin=25mm]{geometry}
\usepackage[skip=6pt,indent=0pt]{parskip}
\usepackage{titlesec}
\usepackage{needspace,placeins}
\titlespacing*{\section}{0pt}{12pt}{6pt}
\titlespacing*{\subsection}{0pt}{10pt}{5pt}
\titlespacing*{\subsubsection}{0pt}{8pt}{4pt}
\titlespacing*{\paragraph}{0pt}{6pt}{0.7em}
\usepackage{amsmath,amssymb,booktabs,tabularx,array}
\usepackage{graphicx,xcolor,microtype}
\usepackage{caption}
\usepackage{float}
\usepackage{algorithm}
\usepackage{algpseudocode}
\usepackage{flafter}

\usepackage[numbers,sort&compress]{natbib}
\usepackage{tikz}
\usetikzlibrary{arrows.meta,positioning,fit,backgrounds,calc}
\usepackage{enumitem}
\usepackage{listings}
\usepackage[colorlinks=true,linkcolor=blue!55!black,citecolor=teal!65!black,urlcolor=blue!55!black]{hyperref}
\definecolor{memorange}{HTML}{C97B37}
\definecolor{memteal}{HTML}{268F89}
\definecolor{mempurple}{HTML}{8564B0}
\definecolor{memblue}{HTML}{4479B3}
\newcommand{\betamem}{\textbf{BetaMem}}
\newcommand{\tool}[1]{\texttt{\small #1}}
\newcolumntype{Y}{>{\raggedright\arraybackslash}X}
\setlist{nosep,leftmargin=*}
\tikzset{box/.style={draw,rounded corners=3pt,line width=.8pt,align=center,font=\sffamily\footnotesize,inner sep=6pt},evidence/.style={box,draw=memorange,fill=memorange!7},state/.style={box,draw=memteal,fill=memteal!7},abstractmem/.style={box,draw=mempurple,fill=mempurple!7},access/.style={box,draw=memblue,fill=memblue!7},flow/.style={-{Stealth[length=2mm]},line width=.8pt,draw=black!60}}
\tikzset{card/.style={box,minimum height=14mm,inner sep=8pt,font=\sffamily\small},labeltext/.style={font=\sffamily\scriptsize,text=black!55},panel/.style={draw=black!15,fill=black!1,rounded corners=5pt},softflow/.style={flow,draw=black!40,line width=.65pt}}

\definecolor{brandink}{HTML}{15243D}
\definecolor{brandrule}{HTML}{C8D5E5}
\definecolor{brandslogan}{HTML}{6754B8}
\usepackage{fancyhdr}
\fancypagestyle{plain}{\fancyhf{}\fancyhead[L]{\raisebox{-1.5mm}{\includegraphics[height=6mm]{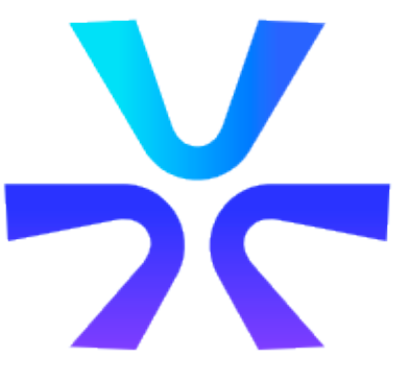}}\hspace{5pt}{\sffamily\bfseries\small\color{brandink}Beta Infinity}}\fancyhead[R]{{\sffamily\bfseries\fontsize{9}{11}\selectfont\color{brandink}Learning What Remains True\ \textcolor{brandslogan}{as the World Changes.}}}\fancyfoot[C]{\sffamily\small\thepage}}
\makeatletter
\renewcommand{\@maketitle}{\newpage\null\vskip 0.4em\begin{center}{\sffamily\bfseries\fontsize{20}{25}\selectfont\color{brandink}\@title\par}\vskip 0.7em\begin{minipage}{\linewidth}\centering\normalsize\normalfont\@author\par\end{minipage}\vskip 0.3em\end{center}\par\vskip 0.4em}
\makeatother

\title{\betamem{}: Learning to Remember a Changing World for Embodied Agents}
\author{\textbf{Mai Liu}\textsuperscript{1}, \textbf{He Sun}\textsuperscript{1,2}, \textbf{Siru Zhong}\textsuperscript{1,3}, \textbf{Kerun Chen}\textsuperscript{1}\\[0.2em]
\textbf{Hao Yan}\textsuperscript{1,4}, \textbf{Xiaohui Lv}\textsuperscript{1}, \textbf{Shuai Tao}\textsuperscript{1}, \textbf{Wulong Liu}\textsuperscript{1}\\[0.35em]
\textsuperscript{1}Beta Infinity\qquad \textsuperscript{2}Peking University\qquad \textsuperscript{4}Beihang University\\[0.15em]
\textsuperscript{3}The Hong Kong University of Science and Technology (Guangzhou)}
\date{}
\begin{document}
\maketitle
\begin{abstract}
Long-horizon embodied agents face temporal confusion and incomplete or distorted records, recurring errors from fixed memory rules, and growing storage and retrieval costs. We present \betamem{}, a framework for building and evolving traceable long-term memory through six linked layers: working, spatial, entity, episodic, experiential, and procedural memory. A dual-level write contract constrains operation executability and source support for content: supervised fine-tuning learns compliant memory operations, while reinforcement learning combines repeated sampling with teacher verification of the original observations to reward complete, accurate records and penalize omissions, distortion, and unsupported content. We formulate failure-driven policy recursive self-improvement through diagnosis along the causal sequence of memory operations and local rule repairs gated by independent replay. Atomic tools frame storage, updates, retrieval, and forgetting as cost--utility allocation under storage and read budgets. Using the open-source Qwen3.5-9B model for memory generation, \betamem{} achieves overall scores of 48.08\% on M3-Bench-robot, 92.08\% on LoCoMo, 72.6\% on OpenEQA EM-EQA, and 62.4\% on EgoLifeQA, the highest among the listed memory methods under their respective evaluation protocols. Training the model for memory generation raises the contract pass rate from 50.9\% to 86.2\% while reducing output tokens per stored action by 47.3\%. A separate Qwen3-VL-8B-Instruct evaluation on EMem-Bench achieves an average success rate of 38.4\% and an average error recurrence rate of 24.9\%. We perform best among evaluated methods using comparably sized open-source models. Category results show that we have advantages in integrating information across observations, recalling events over long histories, and inferring personal habits.
\end{abstract}

\par\noindent\begin{minipage}{\linewidth}
\centering
\captionsetup{hypcap=false,skip=4pt}
\includegraphics[width=\linewidth]{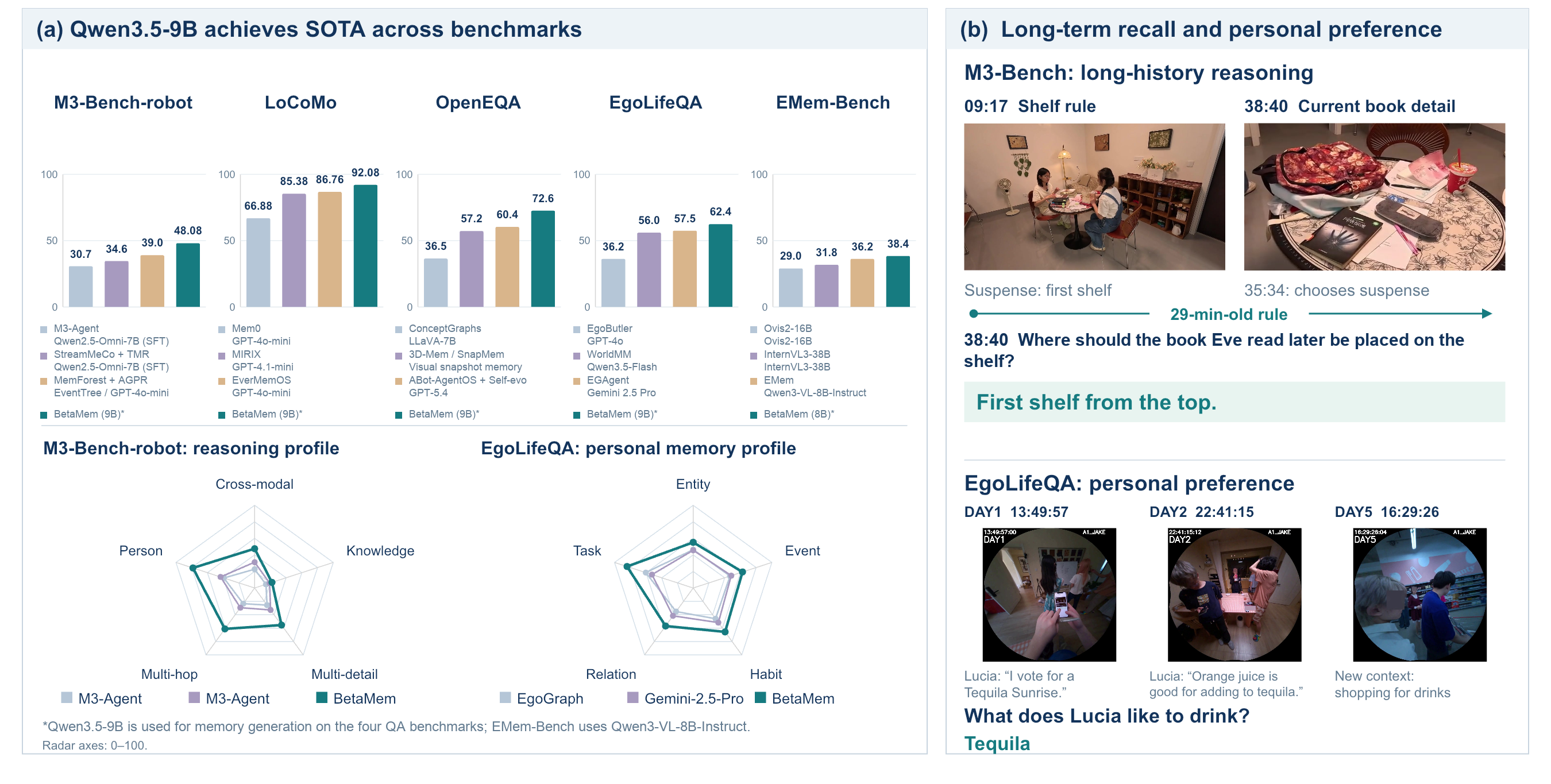}
\captionof{figure}{\textbf{Benchmark performance and evidence-linked memory examples.} (a) Qwen3.5-9B achieves SOTA across benchmarks. Qwen3.5-9B is used for memory generation on the four QA benchmarks; EMem-Bench uses Qwen3-VL-8B-Instruct. (b) M3-Bench links an earlier shelf rule to a later book choice; EgoLifeQA recalls a personal preference across days.}
\label{fig:teaser}
\end{minipage}\par\medskip

\section{Introduction}
\label{sec:introduction}
Embodied agents operating over long horizons must use accumulated experience to answer questions, track a changing environment, and guide subsequent actions~\citep{anwar2024remembr,wang2024karma}. Their observations capture people, objects, places, and events through partial and intermittent sensor streams~\citep{long2025m3agent,lee2026ltmem}. Memory connects these observations across time, providing information beyond the agent's current context~\citep{packer2023memgpt,zhong2026prem}. Its usefulness depends on preserving the temporal context and supporting evidence needed for later tasks~\citep{lee2026ltmem,han2026vista,zhong2026stormbench}.
\begin{figure}[!htbp]
\centering
\includegraphics[width=\linewidth]{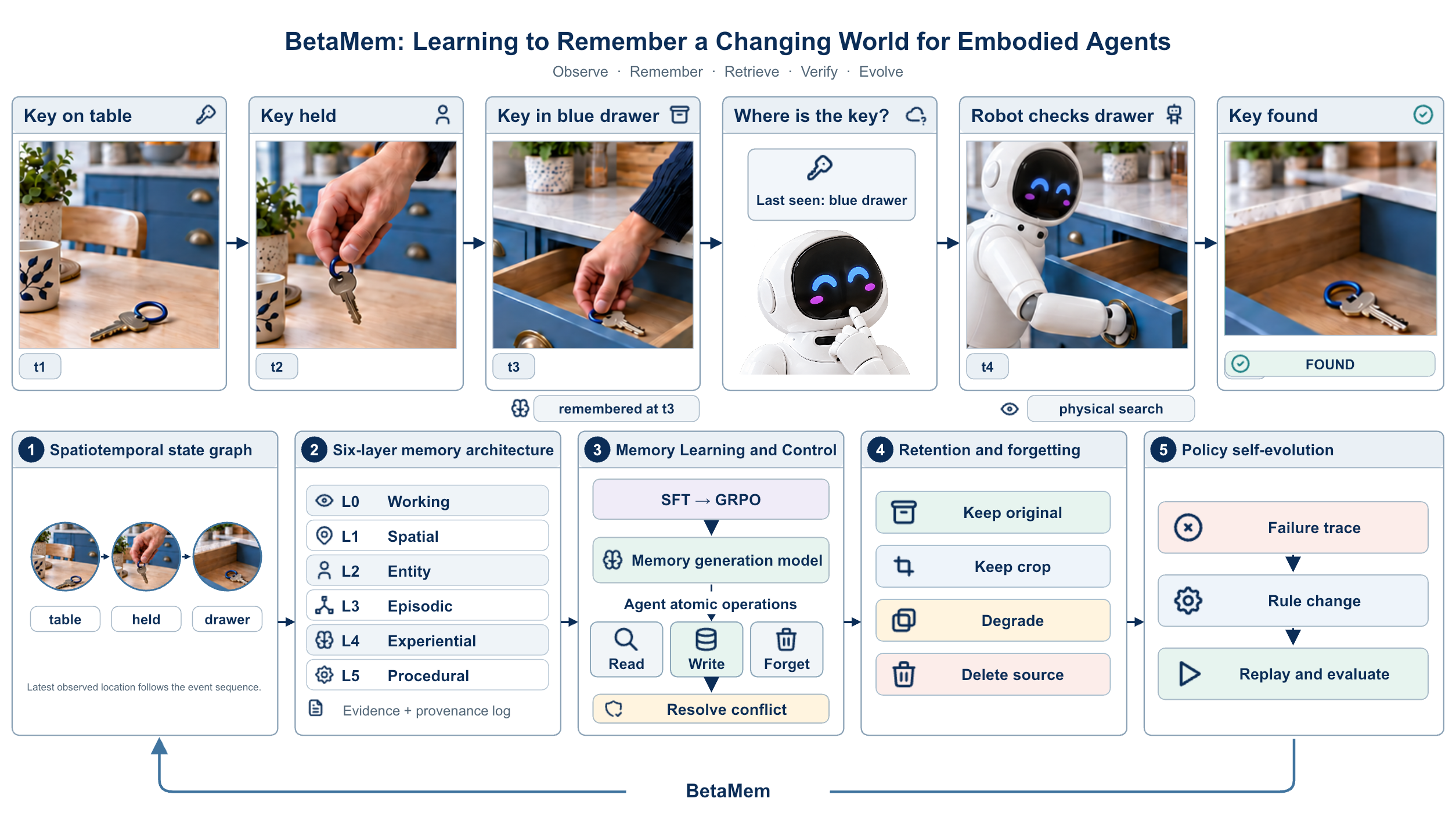}
\caption{\textbf{Evidence-grounded memory with failure-driven policy self-improvement.} In the entryway, the key's observed transitions enter a source-linked spatiotemporal graph and six-layer memory with explicit observation time, validity interval, and revision time. SFT and GRPO train the memory generation model to propose write actions within the Agent's atomic Read/Write/Forget interface, with explicit conflict resolution. Budgeted atomic tools retrieve the $t_3$ drawer placement as a search lead; the robot's new $t_4$ observation verifies the current location and supports a validated state update. Retention and forgetting balance future answerability against source-storage cost. Failure traces localize errors along the memory-operation chain, and local rule repairs undergo independent replay before promotion, closing the loop.}
\label{fig:contributions}
\end{figure}

Recent embodied memory methods organize observations by place, time, and view for later retrieval and exploration~\citep{anwar2024remembr,yang2025threedmem}. Entity-centered episodic and semantic memories connect observations through object and person identities~\citep{long2025m3agent}, while cross-session association preserves identities and competing hypotheses~\citep{lee2026ltmem}. Visual revisiting allows agents to inspect earlier views, regions, and sequences~\citep{han2026vista}. Broader embodied agent systems integrate memory with tool use and verification~\citep{tian2026abotagentos}. These developments make stored experience accessible across extended interactions.

Memory research also addresses how stored experience is maintained and reused. Retention and event compression determine which information remains available for later queries~\citep{zhong2023memorybank,wang2026memforest}. Feedback-driven approaches extract reusable strategies from successful and failed trajectories~\citep{ouyang2026reasoningbank} or search executable harness code through an outer optimization loop~\citep{lee2026metaharness}. Over long interactions, records can become stale, ineffective rules can repeat errors, and observation histories can outgrow storage and read budgets. Memory must therefore adapt its contents and the rules for organizing and retrieving evidence.

A retrieved record must be interpreted in terms of when it was observed, what it supports, and whether its source can still be inspected. For example, an image of a key being placed in a drawer provides a search lead, while confirming its current location requires a new observation (Figure~\ref{fig:contributions}). Maintaining useful, verifiable memory across repeated interactions introduces three coupled challenges:
\begin{itemize}
  \item \textbf{Temporal confusion, omissions, and distortion.} Historical observations can be mistaken for current states, while omitted events or unsupported writes distort entity histories. Memory must preserve each record's temporal scope and source evidence to verify its content and support later answers.
  \Needspace{5\baselineskip}
  \item \textbf{Fixed rules and repeated errors.} Over long task sequences, correcting one record leaves the writing, retention, and retrieval rules unchanged, allowing the same error to recur. The system must use failure feedback to identify memory policies that cause errors, make targeted adjustments, and validate their effectiveness through independent replay before reuse.
  \Needspace{5\baselineskip}
  \item \textbf{Storage and retrieval costs over long horizons.} Continuous sensor streams increase the cost of storing, maintaining, and inspecting memory. Compact summaries can omit details needed by future questions, while retaining every observation consumes substantial storage and read resources. Memory operations must balance their expected answer value against these costs~\citep{wei2026gcr}.
\end{itemize}

To address these challenges, we propose \betamem{}, an evidence-grounded world-memory framework for long-horizon embodied agents. First, the framework organizes observations and experience into six linked memory layers (L0--L5): L0 working memory holds current observations and task state; L1 spatial, L2 entity, and L3 episodic memory link places, people and objects, and event histories; and L4 experiential and L5 procedural memory retain regularities and reusable procedures supported across experiences. The layers record observation time, validity interval, and revision time, while a \emph{dual-level write contract} checks whether update actions are executable and their content is source-supported. During training, the model generates memories from the same observations multiple times; recurring items form candidates for what should be recorded. A teacher model then revisits the original observations to verify support for those candidates. Reinforcement learning uses these checks to reward correct, complete memory updates and penalize omissions, distortion, and unsupported content (Section~\ref{sec:training}). Second, we formulate a failure-driven strategy for \emph{policy recursive self-improvement (RSI)} that follows the causal sequence of memory operations to diagnose failures and gate local rule repairs through independent replay under evidence, permission, and cost constraints. Third, atomic tools expose storage, updates, retrieval, retention, and forgetting. We model allocation as a \emph{cost--utility trade-off} under structured-record, source-data, and read budgets. Together, these mechanisms support long-horizon decisions through traceable evidence, policy adaptation, and budgeted memory use.

Our contributions are as follows:
\begin{itemize}
  \item \textbf{Evidence-grounded world memory for long-horizon embodied agents.} We propose \betamem{} with six linked memory layers (L0--L5) to preserve time-scoped entity and event records with links to source observations. The framework connects memory formation, use, and feedback so that both stored content and memory policies can be revised as experience accumulates.
  \item \textbf{Evidence-constrained learning and adaptation of memory.} We supervise memory writes through checks of action executability and source support, and formulate feedback-driven revision of memory policies. Temporal validity, independent replay, and cost--utility allocation constrain how memory content and policies are updated under storage and read budgets (Sections~\ref{sec:representation}, \ref{sec:training}, \ref{sec:evolution}, and~\ref{sec:tools}).
  \item \textbf{Strong benchmark performance with efficient memory generation using a 9B model.} Using the native open-source Qwen3.5-9B model for memory generation, \betamem{} leads the listed overall comparisons on four memory benchmarks, with 48.08\% overall on M3-Bench-robot, 92.08\% LLM-Judge accuracy on LoCoMo (four non-adversarial categories), 72.6\% LLM-Match on OpenEQA EM-EQA, and 62.4\% exact-match accuracy on EgoLifeQA (A1-JAKE). Training raises the contract pass rate from 50.9\% to 86.2\% while reducing output tokens per stored action by 47.3\%. Analyses highlight advantages in reasoning across observations, adversarial question answering, and person-linked habit recall; a separate Qwen3-VL-8B-Instruct evaluation on EMem-Bench achieves 38.4\% average success rate and 24.9\% average error recurrence rate (Section~\ref{sec:experiments}).
\end{itemize}

\begin{table}[!htbp]
\centering
\begingroup
\newcommand{\capYes}{\textcolor{memteal}{\ensuremath{\checkmark}}}
\newcommand{\capPartial}{\ensuremath{\circ}}
\setlength{\tabcolsep}{4pt}
\renewcommand{\arraystretch}{1.16}
\small
\caption{Mechanism coverage of representative memory and adaptation methods. $\checkmark$: specified; $\circ$: partial or narrower coverage; --: not reported for the defined capability. Coverage refers to the described method design.}
\label{tab:related-capabilities}
\begin{tabularx}{\linewidth}{@{}lY*{6}{c}@{}}
\toprule
\textbf{Method} & \textbf{Main mechanism} & \textbf{T} & \textbf{E} & \textbf{V} & \textbf{L} & \textbf{A} & \textbf{B} \\
\midrule
M3-Agent~\citep{long2025m3agent} & Entity-linked episodic/semantic graph & \capPartial & \capYes & \capPartial & \capYes & -- & \capPartial \\
LT-Mem~\citep{lee2026ltmem} & Volatility-aware state histories & \capYes & -- & \capYes & -- & -- & \capPartial \\
3D-Mem~\citep{yang2025threedmem} & Visual snapshots and learned prefiltering & -- & \capYes & -- & \capYes & -- & \capPartial \\
Embodied-RAG~\citep{xie2024embodiedrag} & Hierarchical semantic-forest retrieval & -- & \capPartial & -- & -- & -- & \capPartial \\
Memory-R1~\citep{yan2025memoryr1} & RL for memory operations and use & \capPartial & -- & -- & \capYes & -- & \capPartial \\
ReasoningBank~\citep{ouyang2026reasoningbank} & Strategies from success and failure & -- & -- & -- & -- & \capPartial & \capPartial \\
KARMA~\citep{wang2024karma} & Scene graph and short-term object states & \capPartial & \capPartial & -- & -- & -- & \capPartial \\
\midrule
\betamem{} & Evidence-grounded world memory & \capYes & \capYes & \capYes & \capYes & \capYes & \capYes \\
\bottomrule
\end{tabularx}
\par\vspace{4pt}
\begin{minipage}{\linewidth}
\footnotesize
\textbf{T}: entity-linked histories that distinguish past and current states; timestamps or unversioned updates alone are partial.
\textbf{E}: on-demand inspection of retained original visual/audio observations; selected identity crops or features are partial.
\textbf{V}: evidence-consistency checks before shared-state updates; conflict voting alone is partial.
\textbf{L}: trained memory construction, management, or retrieval policies, excluding prompt-only experience accumulation.
\textbf{A}: feedback-driven revision of memory or harness rules with separate-episode evaluation; strategy extraction without such a gate is partial.
\textbf{B}: joint constraints on structured storage, source-media storage, and reads; context limits, top-$k$ retrieval, or efficiency controls alone are partial.
\end{minipage}
\endgroup
\end{table}

\section{Related Work}
\label{sec:related}
\subsection{Embodied Memory}
Embodied memory methods organize observations by time, location, and entity identity for later retrieval. ReMEmbR combines descriptions, times, and locations for later agentic retrieval~\citep{anwar2024remembr}; Embodied-RAG indexes embodied descriptions in a multiresolution semantic forest~\citep{xie2024embodiedrag}. KARMA combines a 3D scene graph with short-term object changes~\citep{wang2024karma}, while ConceptGraphs builds open-vocabulary scene graphs from associated views~\citep{gu2023conceptgraphs}. 3D-Mem preserves multiview scene context in memory snapshots and uses frontier snapshots for exploration~\citep{yang2025threedmem}. Entity-centered methods connect observations across time: M3-Agent retrieves episodic and semantic records around entities~\citep{long2025m3agent}, while LT-Mem aligns sessions, associates identities, and preserves uncertainty and competing hypotheses~\citep{lee2026ltmem}. VISTA lets an agent inspect earlier views, regions, and sequences~\citep{han2026vista}. ABot-AgentOS integrates memory with agent capabilities within a broader embodied agent system~\citep{tian2026abotagentos}. \betamem{} organizes current working state (L0), spatial, entity, and episodic records (L1--L3), and experiential regularities and procedural routines derived across experiences (L4--L5) within six linked memory layers. Shared identifiers connect the layers, which combine time-scoped entity states, explicit source-availability information, and evidence checks before shared-memory updates. The embodied agent uses this information to select exploratory actions.

\subsection{General Agent Memory}
Research on agent memory examines how experience is stored, retrieved, and reused across interactions. MemGPT exposes operations across memory tiers~\citep{packer2023memgpt}, MemoryBank studies retention and forgetting~\citep{zhong2023memorybank}, and MemForest compresses events~\citep{wang2026memforest}. PRO-LONG allows a coding agent to search a complete structured interaction log~\citep{fox2026prolong}. Mem-$\alpha$ learns memory construction through reinforcement learning~\citep{wang2025memalpha}, while Memory-R1 trains memory management and use~\citep{yan2025memoryr1}. MemAgent learns to update memory while processing text segments~\citep{yu2025memagent}, and MEM1 jointly trains memory consolidation and reasoning for long-horizon tasks~\citep{zhou2025mem1}. Feedback-driven approaches use past experience to guide future behavior: ReasoningBank extracts reusable strategies from successful and failed trajectories~\citep{ouyang2026reasoningbank}, while Meta-Harness uses an outer loop to search executable harness code~\citep{lee2026metaharness}. Long-horizon embodied memory must manage original images, depth frames, and audio alongside compact event records under storage and read budgets. \betamem{} therefore separates an append-only event and revision log from a finite source-data archive, tracks the temporal scope and source availability of each record, and exposes storage and evidence access through budgeted atomic operations. We further formulate failure-driven memory-policy evolution that diagnoses the failed operation and tests local rule repairs through independent replay under evidence, temporal, and cost constraints.

Table~\ref{tab:related-capabilities} summarizes differences among representative memory systems in temporal state, source-evidence inspection, validation before updates, memory learning, policy adaptation, and resource budgets. Figure~\ref{fig:comparison} shows how temporal ordering and live verification guide a search and validate the resulting state update.

\section{Problem Formulation}
\label{sec:problem}
Let $o_t=(x_t^{\mathrm{rgb}},x_t^{\mathrm{depth}},x_t^{\mathrm{audio}},p_t,a_t)$ denote an observation at time $t$, where $x_t^{\mathrm{rgb}}$, $x_t^{\mathrm{depth}}$, and $x_t^{\mathrm{audio}}$ represent RGB imagery, depth, and audio, respectively. The variables $p_t$ and $a_t$ denote the embodied agent's pose and action metadata. Modalities are optional; retained sources include capture times, availability, and relevant calibration metadata. Let $M_t$ denote the memory state and $U$ its update rule, with $M_{t+1}=U(M_t,o_t)$. For a query $q$, the agent retrieves time-scoped records, source references, evidence availability, and uncertainty. Retrieved evidence guides actions, whose observed outcomes update memory.

\paragraph{Budgeted and evidence-grounded objective.}
Let $B_s$ and $B_e$ bound structured-record and source-data storage, and let $B_r=(N_{\mathrm{calls}},N_{\mathrm{tokens}},T_{\mathrm{latency}})$ bound query-time calls, tokens, and latency. Storage accounting includes metadata, indexes, thumbnails, replicas, and audit records. We optimize $U$, source-retention policy $R$, and retrieval and evidence-review policy $\pi$ to maximize supported answer quality:
\begin{equation}
\max_{U,\pi,R}\;\mathbb E_{q\sim D}[\operatorname{SupportedQuality}(q,M,\pi)]
\quad\text{s.t.}\quad S_s\le B_s,\;S_e\le B_e,\;C(\pi,q)\preceq B_r,
\end{equation}
Here $D$ is the distribution of future queries, $M$ is the memory induced by $U$ and $R$, and $S_s$ and $S_e$ measure the corresponding storage use. $C(\pi,q)$ records the calls, tokens, and latency used to answer $q$, with $\preceq$ applied componentwise. $\operatorname{SupportedQuality}$ is a conceptual objective covering correctness, temporal validity, and source support; experiments use benchmark-specific metrics.

\begin{figure}[!htbp]
\centering
\includegraphics[width=\linewidth]{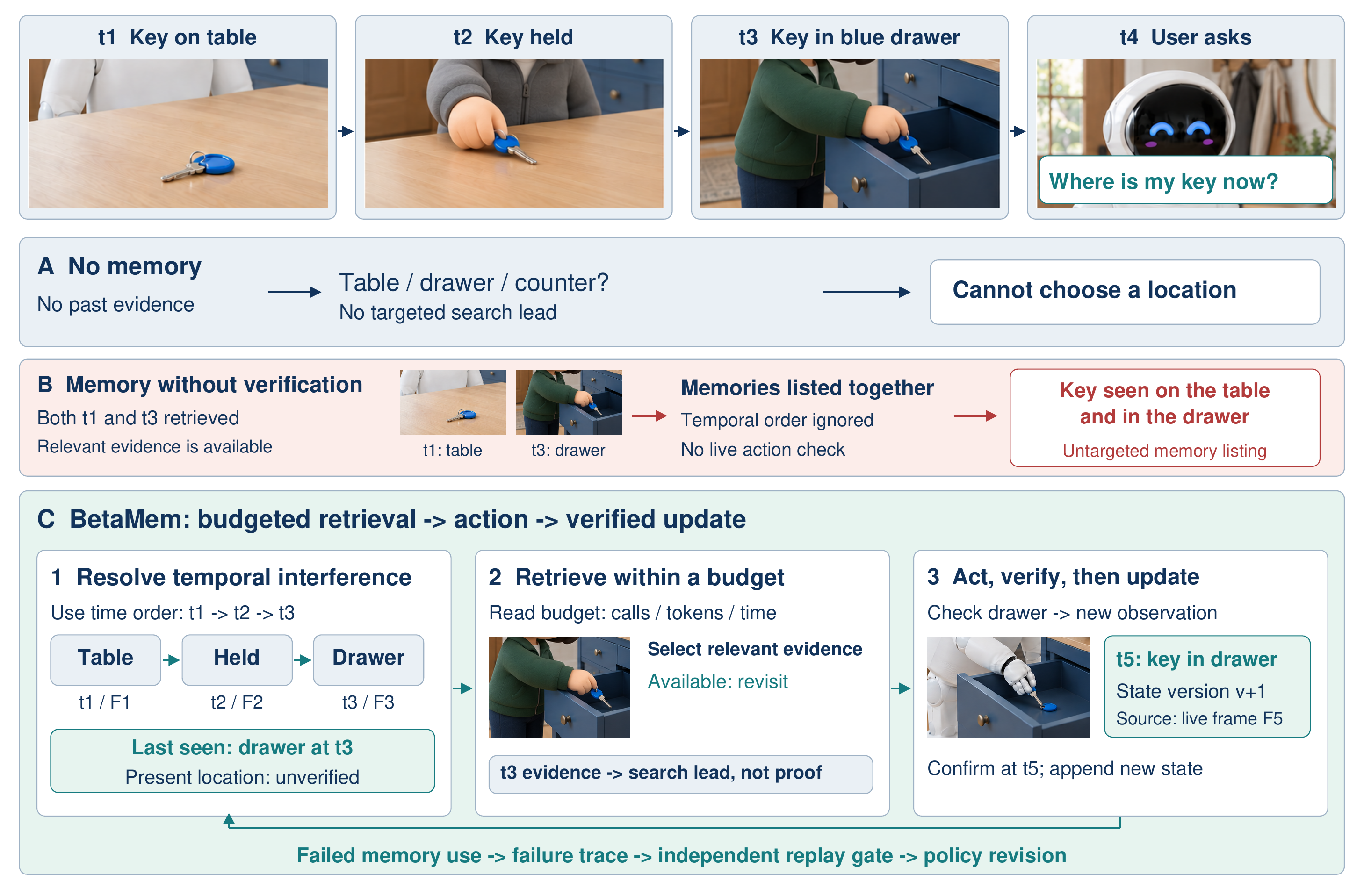}
\caption{Illustrative key search in an entryway under three information-use conditions. (A) Without memory, the agent lacks a targeted search lead. (B) Retrieved $t_1$ and $t_3$ frames are listed without temporal ordering or current-location verification. (C) \betamem{} uses temporal state and budgeted evidence retrieval to select a search lead, checks the drawer, and updates the state after the $t_5$ observation confirms the current location.}
\label{fig:key-search-merged}
\label{fig:comparison}
\label{fig:architecture}
\end{figure}

Answers preserve the temporal scope and availability of their evidence. Historical records guide search; current-state confirmation requires new observations (Figure~\ref{fig:comparison}).

\section{\betamem{}: Evidence-Grounded World Memory}
\label{sec:method}
\betamem{} organizes long-horizon memory around three coupled decisions: which observations support a record, how failures should change memory policy, and where limited storage and read resources should be allocated. The online pipeline converts observations into time-scoped, source-linked records and retrieves them to support subsequent decisions. Layered representation and dual-level write learning address temporal confusion, omissions, and distortion (Sections~\ref{sec:representation} and~\ref{sec:training}); failure-driven policy RSI adapts the rules governing memory use (Section~\ref{sec:evolution}); atomic operations expose the cost--utility trade-off in retaining and accessing evidence (Section~\ref{sec:tools}).
\subsection{Temporal and Evidence-Grounded Memory}
\label{sec:representation}
\betamem{} represents observation histories through a temporal state graph, an append-only event log, and a finite source-data archive. The graph indexes supported entity states, the log preserves the transitions and revisions that produced them, and the archive retains selected original observations. Shared entity, event, and source identifiers connect these structures. The source evidence index links source identifiers to original observations and their capture metadata; access permissions and availability are checked when that observation is requested.

\paragraph{State records with spatiotemporal context.}
A state record specifies its content, spatial and temporal context, and supporting evidence:
\begin{equation}
\begin{aligned}
s&=(e,f,v,\mathcal{S},\mathcal{T},c,E,\nu,\sigma),\\
\mathcal{T}&=(I_{\mathrm{valid}},t_{\mathrm{obs}},t_{\mathrm{commit}}).
\end{aligned}
\end{equation}
where $e$ denotes an entity, $f$ a field or relation, and $v$ its value. Spatial context $\mathcal{S}$ identifies the associated place or container, reference coordinate frame, and map version; when supported by observations, it also records position, pose, and their uncertainty. Spatial context may be empty for non-spatial records, such as preferences. Temporal context $\mathcal{T}$ groups the validity interval $I_{\mathrm{valid}}$, source capture time $t_{\mathrm{obs}}$, and revision commit time $t_{\mathrm{commit}}$. The validity interval specifies the time described by the record. Confidence $c$, source identifiers $E$, revision number $\nu$, and status $\sigma$ complete the record. Status distinguishes observed, inferred, conflicting, stale, and unknown information. Uncertain transition times are represented as intervals, and claims extending beyond direct observational support are marked as inferred. Spatial references locate the state in the environment, while the separate time fields permit corrections without losing capture times.

\begin{figure}[!htbp]
\centering
\includegraphics[width=\linewidth]{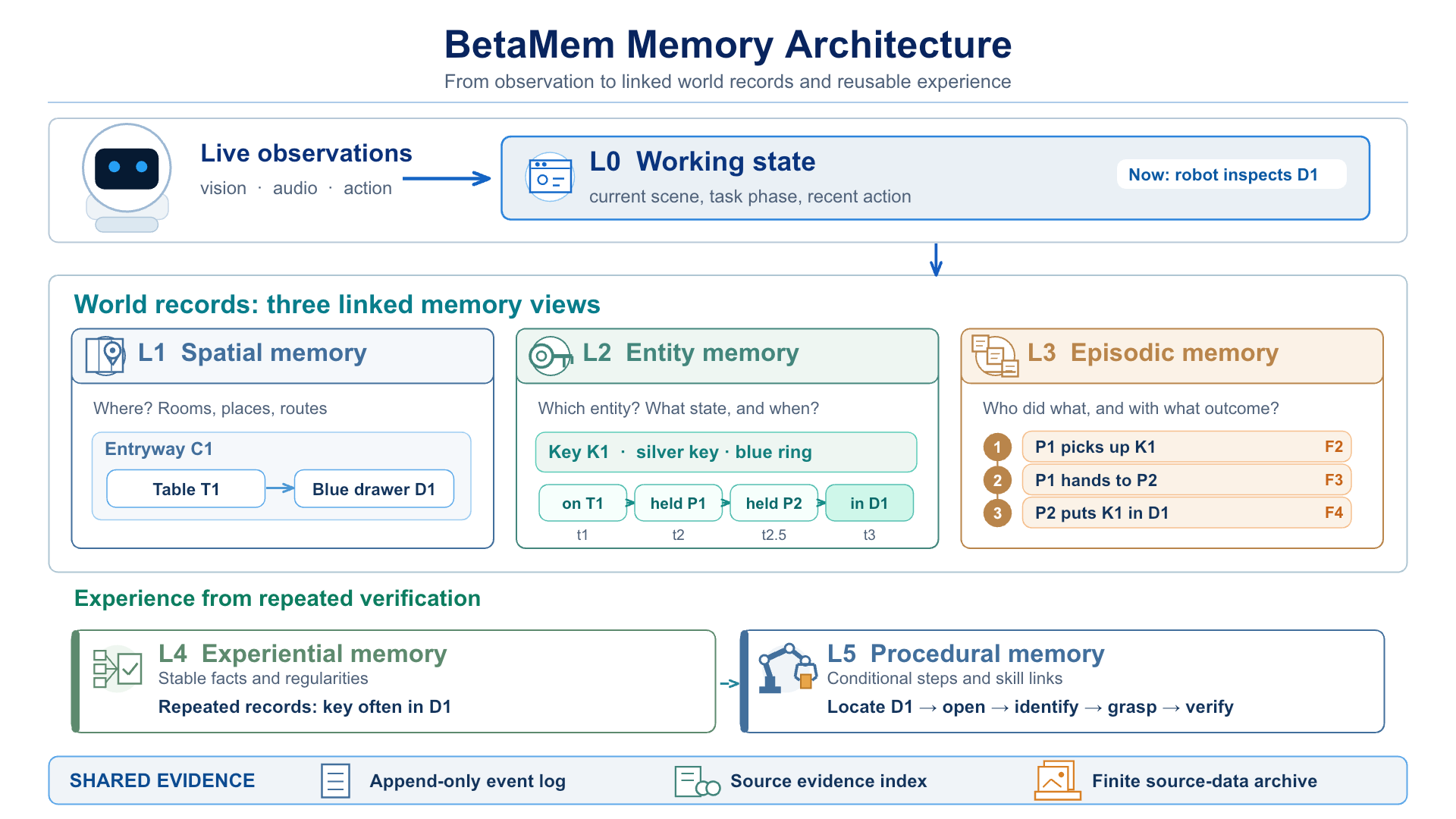}
\caption{Six-layer memory architecture. Current observations (L0) update linked spatial (L1), entity (L2), and episodic (L3) records. Corroborated histories support experiential regularities (L4) and reusable procedures (L5). A shared event log, source evidence index, and finite archive preserve provenance across layers.}
\label{fig:memory_functions}
\end{figure}

\paragraph{Linked memory layers.}
The six layers in Figure~\ref{fig:memory_functions} organize records by their role in perception, world-state tracking, and action. L0 stores the current sensor window, task stage, and action history. Its observations update three parallel, cross-referenced structures: L1 spatial memory stores maps, places, routes, and coordinate frames; L2 entity memory maintains object and person identities with time-scoped attributes and relations; and L3 episodic memory stores event participants, action order, outcomes, and source links. An L3 transition can revise an L2 state, while L1 links that transition to a place and map version. L4 experiential memory consolidates corroborated histories into regularities with explicit scope and exceptions. L5 procedural memory stores task preconditions, place and entity references, callable skills, expected observations, success checks, and recovery branches. Regularities and reusable procedures require support across distinct experiences. During execution, the embodied agent selects skill references from L5 according to the current L0 state; the external controller executes the selected skills and returns observations for subsequent updates.
The layers share one database through linked indexes and views. The event log and source archive span all six layers, preserving provenance across consolidation and revision. Cross-layer retrieval can therefore connect an entity state to its place, reconstruct the events that produced it, and inspect retained source details when needed. These links support both reasoning across observations and the consolidation of person-linked regularities.

\begin{figure}[!htbp]
\centering
\includegraphics[width=\linewidth]{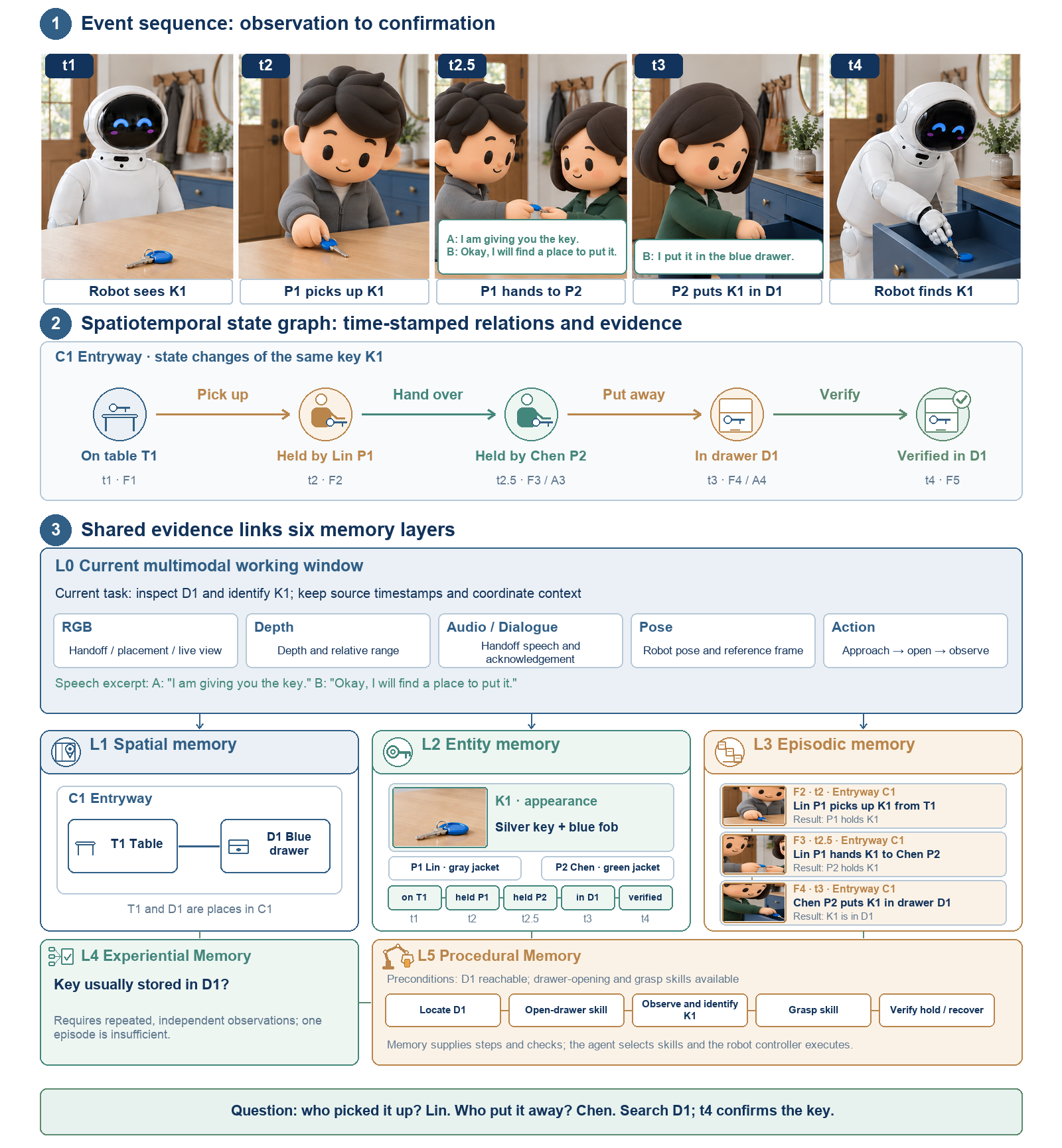}
\caption{Multimodal memory construction in an illustrative key transfer. L0 observations link pickup, handoff, and placement to distinct actors and time-stamped image, audio, depth, pose, and action sources. The spatiotemporal state graph connects people, the key, events, and places in one visual path, retaining timestamps and source markers for each state. L1--L3 preserve spatial context, identity, and event order; corroborated episodes support L4 regularities and L5 procedures. The $t_4$ observation confirms the current location.}
\label{fig:key_event_example}
\end{figure}

\paragraph{Spatiotemporal state graph and multimodal event association.}
The spatiotemporal state graph links people, objects, places, and events, retaining spatial references, times, and sources for entity states and their transitions.
Visual, audio, depth, pose, and action records retain their capture times, coordinate context, and source identifiers. Transcripts additionally retain speaker attribution and links to their original audio. Entity identifiers and sensor-source identifiers use separate namespaces. This association preserves the distinction among spoken intentions, reported outcomes, and directly observed state changes. Figure~\ref{fig:key_event_example} illustrates this structure with a key transfer. In entryway $C1$, $P1$ picks up key $K1$ from table $T1$ and hands it to $P2$, who places it in the entryway cabinet drawer $D1$. Separate pickup, handoff, and placement events retain their actors and linked image and audio sources, allowing retrieval to distinguish who acquired the key from who stored it. The spatiotemporal state graph presents one visual path for the same key in entryway $C1$: on table $T1$, held by Lin $P1$, held by Chen $P2$, placed in drawer $D1$, and finally verified by a new observation. Pickup, handoff, placement, and verification connect people, the object, and space; each state retains its time and image or audio source markers. Repeated corroborated episodes can subsequently support an L4 habit or an L5 search procedure.

\paragraph{Incremental construction and validated revision.}
\label{sec:formation}
Each observation first receives a source record containing its capture time and sensor metadata. Windows are segmented at scene changes, object interactions, action boundaries, or a maximum duration. The memory generation model maps each window and its available history to typed entity, relation, and event proposals with source references. Association checks resolve proposals against existing identities and places using appearance, category, position, and temporal continuity. Ambiguous matches retain competing hypotheses. Pose and geometry preserve their coordinate frames and uncertainty; spatial revisions require a consistent map alignment. This association builds on cross-session entity tracking~\citep{lee2026ltmem} and spatial scene graphs~\citep{gu2023conceptgraphs}. Before commitment, each proposal is checked against its source evidence, temporal scope, and the versions of the records it uses. An accepted transition appends an event and revises the relevant state record while preserving its history. Episode summaries retain links to the underlying events and observations; exact times, positions, uncertainty, and conflicts remain in the structured records. Consolidation into L4 or L5 requires corroboration across distinct episodes. Write learning supervises update executability and source-supported content.

\subsection{Dual-Level Memory Write Learning}
\label{sec:training}
Let $w$ denote an observation window with its available memory history, and let $a$ denote a candidate list of memory-write actions. The memory generation model learns to map $w$ to executable, source-supported updates of the layered memory. A \emph{dual-level write contract} separates interface validity from content validity: interface checks assess action format, fields, identifiers, and value types, while content checks assess source support, coverage, temporal consistency, and cross-layer links. This separation provides direct supervision for malformed actions, omitted observations, and unsupported writes.

\paragraph{Training sample construction and data verification.}
\label{sec:training-data}
A frontier teacher processes observation windows through the streaming service to generate candidate actions. Each supervised example pairs a window and its available history with a checked action list. A deterministic checker applies 26 traceable rules, repairing or removing recoverable contract violations and controlling the distribution of target actions. These examples establish the interface-level supervision used for supervised fine-tuning (SFT). Content-level supervision additionally requires targets for observations that the model may omit. After SFT, the model processes each training video independently $K{=}4$ times. Items recorded in at least two runs form the approximate \emph{should-record set} $V_w$ for window $w$; single-run items are excluded. Vote counts measure recording consistency. A low count can arise from either an omission or an uncertain observation, so source verification determines eligibility for additional coverage weight. We use \emph{teacher replay} with \texttt{qwen3.7-max} to inspect a candidate's original frames and available history before GRPO training. It checks entity identity, attributes, temporal correspondence, state changes, and event participants, order, and outcomes, returning the supporting source frames. Deterministic checks verify the returned frame references and timestamps. Verified low-vote items receive a bounded omission-compensation boost that decreases from two to three to four votes; normalization preserves the mean weighting level. Unclear items receive no boost, and contradicted items are excluded from coverage targets. Missing detections are treated as uncertain evidence. Replay uses observations available at the window, excluding future frames and evaluation answers. Verification covers both entity and episodic memory. For entity memory, it checks observational support for identities, attributes, and spatial relations; for episodic memory, it checks source consistency of participants, action order, state transitions, and outcomes. Both types follow the same treatment: source-verified low-vote items receive a modest boost, items with insufficient evidence receive no boost, and items contradicted by observations are excluded from coverage targets. A verified four-vote item retains its base weight. Thus, votes screen candidates and measure recording stability, while observations determine factual support and the training decision.

\paragraph{Contract-based reinforcement objective.}
\label{sec:training-objective}
We score candidate writes against their observations, linked history, and coverage targets. Direct write-level feedback separates memory construction from downstream retrieval and answer generation. The objective first gates parseability, then combines target coverage with penalties for unsupported content and invalid or redundant actions. Candidate actions are scored before repair. Coverage requires recording the observed attribute or transition with a consistent entity link; copying an entity name alone earns no coverage for that transition. For a window $w$ and candidate action list $a$, the content score and reward are
\begin{align}
s(w,a) =\;&
1 + \lambda_{\mathrm{cov}}\,\mathrm{cov}_{0:5}(w,a) \nonumber\\
&- \lambda_{\mathrm{omit}}\,\mathbf{1}[\mathrm{cov}_{0:5}{=}0 \wedge T_w{\ne}\emptyset]
- \sum_{\ell=0}^{5}\beta_\ell E_\ell(w,a) \nonumber\\
&- \lambda_{\mathrm{dup}}\,\min(n_{\mathrm{dup}},\kappa_{\mathrm{dup}})
- \lambda_{\mathrm{red}}\,\min(n_{\mathrm{red}},\kappa_{\mathrm{red}}) \nonumber\\
&- \lambda_{\mathrm{act}}\,(n_{\mathrm{act}}{-}\kappa_{\mathrm{act}})^{+}
- \lambda_{\mathrm{oid}}\,\min(n_{\mathrm{oid}},\kappa_{\mathrm{oid}}) \nonumber\\
&- \lambda_{\mathrm{fld}}\,\mathbf{1}[\mathrm{flood}]
- \lambda_{\mathrm{ground}}\,\mathbf{1}[\mathrm{ungrounded}]
\label{eq:write-content}
\end{align}
\begin{equation}
r(w,a)=\begin{cases}
-1, & g_{\mathrm{fmt}}(a)=0,\\[2pt]
\varepsilon+(1-\varepsilon)\operatorname{sigmoid}(s(w,a)/\tau),
& g_{\mathrm{fmt}}(a)=1.
\end{cases}
\label{eq:write-reward}
\end{equation}
Here $g_{\mathrm{fmt}}(a)$ is one for a parseable action list and zero otherwise. An unparseable list receives $-1$; parseable lists are mapped into $(\varepsilon,1)$, where $0<\varepsilon<1$, $\tau>0$, and $\operatorname{sigmoid}(z)=1/(1+e^{-z})$. This ordering separates unusable actions from recoverable content errors while retaining graded feedback within the parseable group. The positive coefficient $\lambda_{\mathrm{ground}}$ penalizes insufficient spatial grounding additively. Scale selection and numerical-saturation checks use development data.

\paragraph{Coverage and layer consistency.}
The coverage term is
$\mathrm{cov}_{0:5}(w,a)=\frac{\sum_{\ell=0}^{5}m_{w\ell}\alpha_\ell C_\ell(w,a)}{\sum_{\ell=0}^{5}m_{w\ell}\alpha_\ell}$,
where $C_\ell\in[0,1]$ is the correctly covered fraction of target set $T_{w\ell}$, $m_{w\ell}\in\{0,1\}$ indicates whether layer $\ell$ participates, and $\alpha_\ell>0$ is its weight. Coverage is zero when the denominator is zero. Layer-wise normalization balances entity, event, regularity, and procedure targets. Entity targets retain the vote weights of $V_w$ and require agreement with current detections and state. With $T_w=\bigcup_{\ell=0}^{5}T_{w\ell}$, a nonempty target set with zero coverage incurs the complete-omission penalty. The layer error $E_\ell(w,a)\geq0$ counts unsupported writes or incorrect links, with weight $\beta_\ell>0$. L0 checks temporal alignment of observations, task stage, and action history; L1 checks places, spatial relations, and reference frames; L2 checks identity, attributes, and validity intervals; and L3 checks event participants, order, and provenance. L4 regularities and L5 procedures are checked against corroborating linked history, including scope, exceptions, preconditions, skill steps, expected observations, and verified outcomes. A single event does not supply coverage for an unsupported regularity or unvalidated procedure.

\paragraph{Action-level penalties.}
$n_{\mathrm{dup}}$ counts creation of stored names absent from current detections; $n_{\mathrm{red}}$ counts location updates for items absent from the detection segment; $n_{\mathrm{act}}$ counts actions; and $n_{\mathrm{oid}}$ counts repeated entity identifiers. The flood indicator identifies runs of redundant updates, and the ungrounded indicator identifies lists whose new boxes all miss the detected boxes. Positive coefficients $\lambda$ set reward and penalty strengths, $\kappa$ caps counts or the allowed action count, and $(x)^+=\max(x,0)$. Each error is assigned to one penalty category to avoid double counting between layer and action errors. Weights are fixed within a training run; coverage and penalty components are logged separately. Synthetic windows with known targets support checks of reward behavior for correct updates, omissions, fabricated entities, and copied names. Terminal rewards have been used for memory operations in text settings~\citep{wang2025memalpha,yan2025memoryr1,yu2025memagent,zhou2025mem1}, while the model for egocentric memory generation of \citet{xiong2026memlife} uses frozen evaluators and QA pairs. Our formulation supplies feedback at the write decision through deterministic coverage and error checks. DAPO~\citep{yu2025dapo} uses sampling for zero-variance groups; our penalties distinguish recoverable violations within groups.

\paragraph{Two-stage optimization.}
\label{sec:training-stages}
SFT initializes the 9B memory generation model from checked teacher examples, learning the action format and the content retained in the training dataset. Window-level Group Relative Policy Optimization (GRPO)~\citep{shao2024deepseekmath} then samples $G{=}8$ alternative action lists for the same window and uses their relative rewards to update the model. Holding the observation and history fixed isolates differences among write decisions. A KL constraint to the SFT-initialized model limits policy drift. Training configuration and operational measurements are reported in Section~\ref{sec:training-effects}. These stages establish the initial write policy; the failure-driven mechanism in Section~\ref{sec:evolution} addresses subsequent errors in memory-use rules.

\subsection{Failure-Driven Memory Policy RSI}
\label{sec:evolution}
\begin{figure}[!htbp]
\centering
\includegraphics[width=\linewidth]{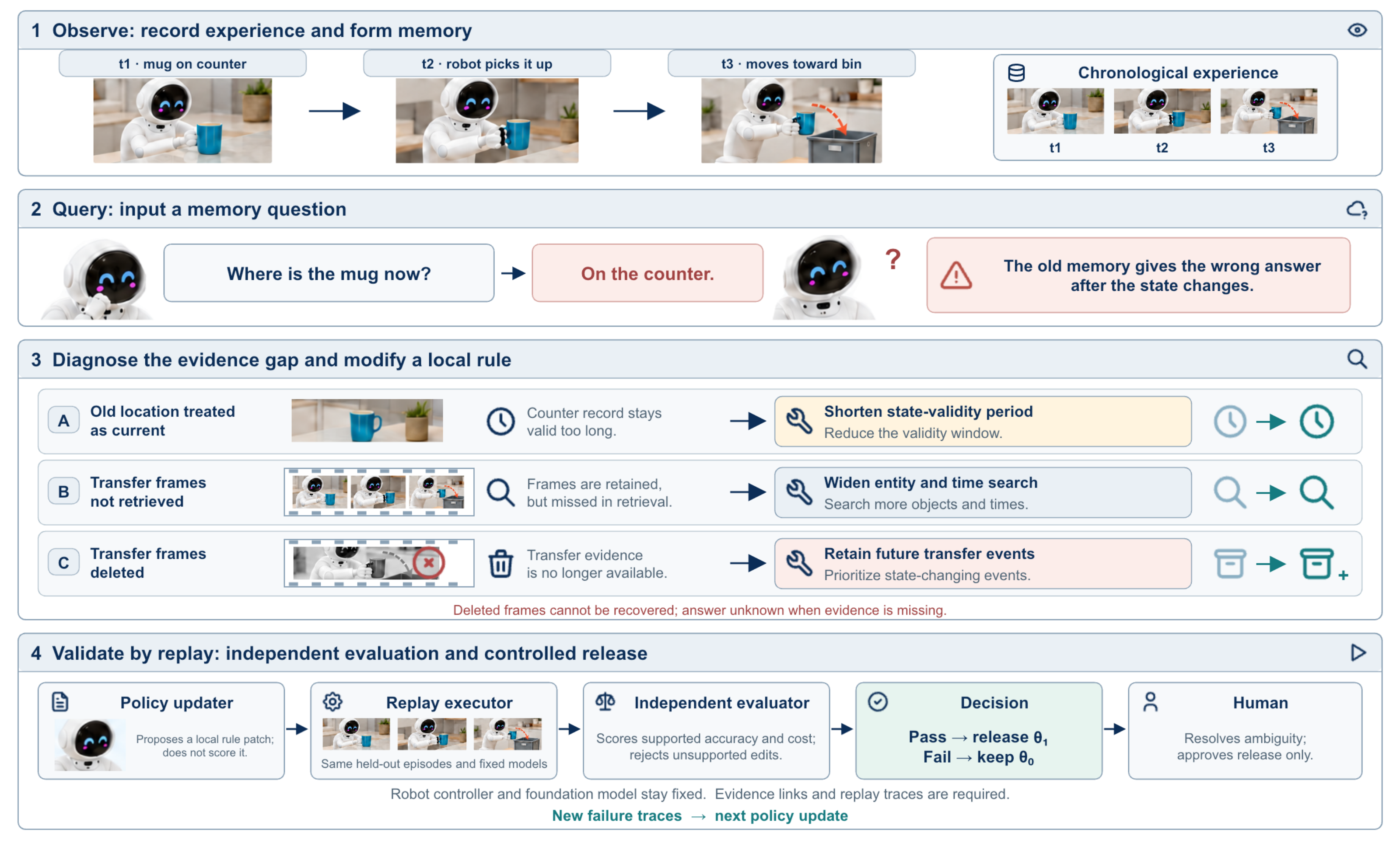}
\caption{Illustrative failure-driven policy RSI. Failure traces identify errors in state validity, retrieval, or retention. An LLM-based updater proposes local repairs for independent replay against the active policy on held-out episodes with fixed models. Evidence and cost checks gate promotion; subsequent failures initiate another update.}
\label{fig:self_evolution}
\end{figure}

We formulate policy recursive self-improvement (RSI) as a sequence of diagnosed failures, local rule repairs, and independently validated policy versions. Let $\theta=(\theta_{\rm write},\theta_{\rm retain},\theta_{\rm retrieve},\theta_{\rm review},\theta_{\rm fresh})$ denote the rules governing writing, source retention, retrieval, evidence review, and state validity. The mechanism updates these rules while keeping foundation-model weights and the robot controller fixed. Each accepted policy governs subsequent episodes, whose failures can trigger further updates (Figure~\ref{fig:self_evolution}).

\paragraph{Failure diagnosis.}
A failure trace $\tau$ links a memory-dependent answer or decision to verified feedback, observation and event identifiers, tool calls, retrieved items, source availability, policy version, and resource use. Feedback distinguishes human correction, direct observation, and inferred task outcome by reliability. The diagnostic process follows the dependencies among observation capture, retention, retrieval, review, and state use. It first establishes whether the relevant observation was captured and retained, then checks whether retrieval found it, review inspected it, and the state-validity rule used it correctly. Missing observations provide no evidence for a memory-rule repair; ambiguous identity or outcome feedback remains unresolved. An LLM-based updater analyzes the failure trace to identify memory rules that may have contributed to the failure and provides evidence supporting this attribution. The proposal records the affected decision, missing or misused evidence, and episodes on which a repair should matter. Attribution remains a hypothesis until replay tests the candidate; repeated queries from one episode are grouped when assessing independent failures. This limits repair to the implicated operation.

\paragraph{Constrained local repair.}
Given active policy $\theta_0$ and diagnosed rule $d$, the updater constructs candidates in a local neighborhood $N(\theta_0,d)$ that changes that rule. Each candidate retains its parent version, target failure class, motivating traces, affected cases, and predicted cost. Candidates are selected by evidence-supported answer quality on an independent regression set:
\begin{equation}
\begin{aligned}
\theta^*&=\underset{\theta\in N(\theta_0,d)}{\arg\max}\;\widehat Q_{\rm sup}^{\,\rm val}(\theta)\\
\text{s.t.}\quad \mathbf C^{\rm val}(\theta)&\preceq\mathbf B,\qquad
\widehat U_{\rm rev}^{\,\rm val}(\theta)\leq\widehat U_{\rm rev}^{\,\rm val}(\theta_0)+\epsilon_u .
\end{aligned}
\end{equation}
Here $\theta^*$ is the selected candidate, $\rm val$ denotes the independent set excluded from proposal generation, and hats denote empirical estimates. $\widehat Q_{\rm sup}^{\,\rm val}$ measures supported-answer quality, while $\widehat U_{\rm rev}^{\,\rm val}$ counts unsupported state revisions. The vector $\mathbf C^{\rm val}$ measures structured storage, source-data storage, read calls, tokens, and latency against componentwise limits $\mathbf B$. The tolerance $\epsilon_u$ bounds any increase in unsupported revisions. Scoring rules, tolerances, and the required target-class improvement $\delta$ are fixed before candidate generation. Access and deletion constraints determine candidate eligibility.

\paragraph{Independent replay and version promotion.}
A separate executor replays the active and candidate policies on identical observation sequences and queries with fixed perception, language-model settings, and controller. An independent evaluator measures supported answers, unsupported revisions, abstentions, and resource use. Changes to retrieval or state-validity rules are evaluated using existing memory records and their source evidence. Changes to retention or forgetting rules are evaluated by rebuilding memory from an authorized archive of original observations, so that the effects of discarding evidence can be assessed. Replay proceeds chronologically: each retention decision uses only observations available at that time, without access to later queries. Promotion requires the prespecified target-class gain and satisfaction of evidence, permission, and cost constraints, followed by a staged canary test. The new policy retains its parent and rollback information, and subsequent responses carry the policy version. This produces the recursive sequence $\theta_0\rightarrow\theta_1\rightarrow\theta_2\rightarrow\cdots$, with independent replay at every transition. Algorithm~\ref{alg:policy-evolution} summarizes the cycle; replay returns quality $Q_\theta$, unsupported revisions $U_\theta$, resource use $\mathbf C_\theta$, and target-class gain $\Delta_\theta$ over $\theta_0$.

\begin{algorithm}[!htbp]
\caption{Failure-Driven Memory Policy RSI}
\label{alg:policy-evolution}
\small
\begin{algorithmic}[1]
\Require Trace $\tau$, active policy $\theta_0$, regression set $D_{\rm val}$, budgets $\mathbf B$, tolerance $\epsilon_u$, margin $\delta$
\Ensure Next policy version $\theta_{\rm new}$
\State $d\gets$ diagnose the failed observation, retention, retrieval, review, or validity rule
\If{evidence is insufficient to support the diagnosis}
  \State \Return $\theta_0$
\EndIf
\State $\Theta\gets N(\theta_0,d)$ \Comment{change only the implicated rule}
\State $(Q_0,U_0,\mathbf C_0)\gets$ independently replay the active policy on $D_{\rm val}$
\State $A\gets\emptyset$ \Comment{eligible candidates}
\For{each $\theta\in\Theta$}
  \State $(Q_\theta,U_\theta,\mathbf C_\theta,\Delta_\theta)\gets$ independently replay $\theta$ with the same observations and questions
  \If{access and deletion constraints hold, and $\mathbf C_\theta\preceq\mathbf B$}
    \If{$U_\theta\leq U_0+\epsilon_u$ and $\Delta_\theta\geq\delta$}
      \State $A\gets A\cup\{\theta\}$
    \EndIf
  \EndIf
\EndFor
\If{$A=\emptyset$}
  \State \Return $\theta_0$
\EndIf
\State $\theta^*\gets\arg\max_{\theta\in A}Q_\theta$
\If{$\theta^*$ passes the staged canary test}
  \State Publish a new version; retain its parent, change record, and rollback information
  \State \Return $\theta^*$
\EndIf
\State \Return $\theta_0$
\end{algorithmic}
\end{algorithm}

\subsection{Budgeted Memory Operations}
\label{sec:tools}
Atomic memory calls make the costs and state effects of storage, retrieval, updates, and forgetting explicit. \betamem{} exposes eight interfaces grouped as Read (2), Write (3), Manage (2), and Report (1), all operating on the shared database (Table~\ref{tab:memory_calls}). Calls return temporal scope, provenance, versions, and outcomes. Their grouping identifies operational purpose; permissions are checked separately for each record or source. These interfaces supply the operations over which storage and read budgets are allocated.

\begin{table}[!htbp]\centering\footnotesize
\begin{tabularx}{\linewidth}{@{}p{0.31\linewidth}YY@{}}\toprule
Agent-facing call & Required input and purpose & Return and state effect \\
\midrule
\multicolumn{3}{@{}l@{}}{\textbf{Read (2): retrieve records or inspect sources; no accepted-state change}} \\[2pt]
\tool{read\_memory} & Question or entity, time/place, mode, budget; retrieve memory & Answer or ranked entries, sources, uncertainty; read only \\
\tool{read\_evidence} & Source ID, region, permission, budget; inspect source data & Permitted source data and metadata, or unavailable; read only \\
\midrule
\multicolumn{3}{@{}l@{}}{\textbf{Write (3): append observations, outcomes, or proposals; validate state updates}} \\[2pt]
\tool{write\_observation} & Observation, time, pose, source; record perception & Source record and entity/event candidates \\
\tool{write\_execution} & Task step, action trace, observed result; record execution & Sourced outcome event; no unobserved-state inference \\
\tool{write\_candidate} & Proposed memory, time, source, evidence; propose write & Proposal and validation status; accepted only after checks \\
\midrule
\multicolumn{3}{@{}l@{}}{\textbf{Manage (2): apply retention and conflict rules; preserve audit outcomes}} \\[2pt]
\tool{manage\_retention} & Record or source data, reason, rule, version; request forgetting & Deleted, degraded, expired, or rejected, with dependency outcome \\
\tool{manage\_conflict} & Conflicting entries, sources, versions; request resolution & Merge, select, coexist, or await observation; audit preserved \\
\midrule
\multicolumn{3}{@{}l@{}}{\textbf{Report (1): record issues and traces; no direct memory or policy change}} \\[2pt]
\tool{report\_feedback} & Answer or decision ID, verified outcome or correction, source; flag memory use & Intake status and trace ID; no direct memory or policy change \\
\bottomrule\end{tabularx}
\caption{Eight agent-facing memory calls grouped by purpose and state effect: Read (2), Write (3), Manage (2), and Report (1). All calls share version and provenance checks.}
\label{tab:memory_calls}
\end{table}

\paragraph{Evidence-availability-aware memory retrieval.}
\tool{read\_memory} accepts a query or entity description, spatiotemporal constraints, a read budget, and a return mode. It matches candidate entities and retrieves linked state, spatial, and event records, using appearance features or text descriptions as additional matching cues when identity is uncertain. In answer mode, it returns an answer or search lead with source references and uncertainty; in record mode, it returns relevance-ranked memory entries for subsequent reasoning. Results preserve each record's validity interval and source-availability status. \tool{read\_evidence} reads original observations or specified regions by source identifier, subject to access permissions and read budgets, and returns source data and capture metadata; inaccessible sources return an unavailable status. The two interfaces separate structured memory retrieval from original-evidence access, allowing the latter's permissions and costs to be controlled independently. Historical memory identifies search targets, while new observations verify the current state; verifiable details depend on source availability.

\paragraph{Memory writing, retention management, and conflict resolution.}
\tool{write\_observation} records sensor observations with timestamps, poses, and sources, producing entity and event candidates for validation. \tool{write\_execution} records task steps, action traces, and observed execution outcomes, linking actions to outcome events. \tool{write\_candidate} submits proposed memory entries with sources, validity intervals, confidence, and supporting evidence. Updates to accepted states must pass the shared validation process in Section~\ref{sec:representation}; outcomes without observational support are not written as confirmed states. \tool{manage\_retention} applies retention and forgetting rules to specified records or source data at expected versions, returning deletion, degradation, expiration, or rejection outcomes together with the disposition of associated dependencies. \tool{manage\_conflict} checks conflicting records, sources, and versions, returning an evidence-supported merge, selection, coexistence, or pending-observation outcome while preserving superseded records and resolution history for auditing.

\paragraph{Diagnostic feedback and policy improvement.}
\tool{report\_feedback} links an answer or memory-use decision to verified execution outcomes, corrections, and source evidence. It returns an intake status and failure-trace identifier for failure diagnosis and rule repair in Policy RSI. This interface records feedback without directly changing memory states or promoting policies. Memory changes use the corresponding write or management interfaces, while policy versions are selected through the independent replay and validation process in Section~\ref{sec:evolution}.

\begin{figure}[!htbp]
\centering
\includegraphics[width=\linewidth]{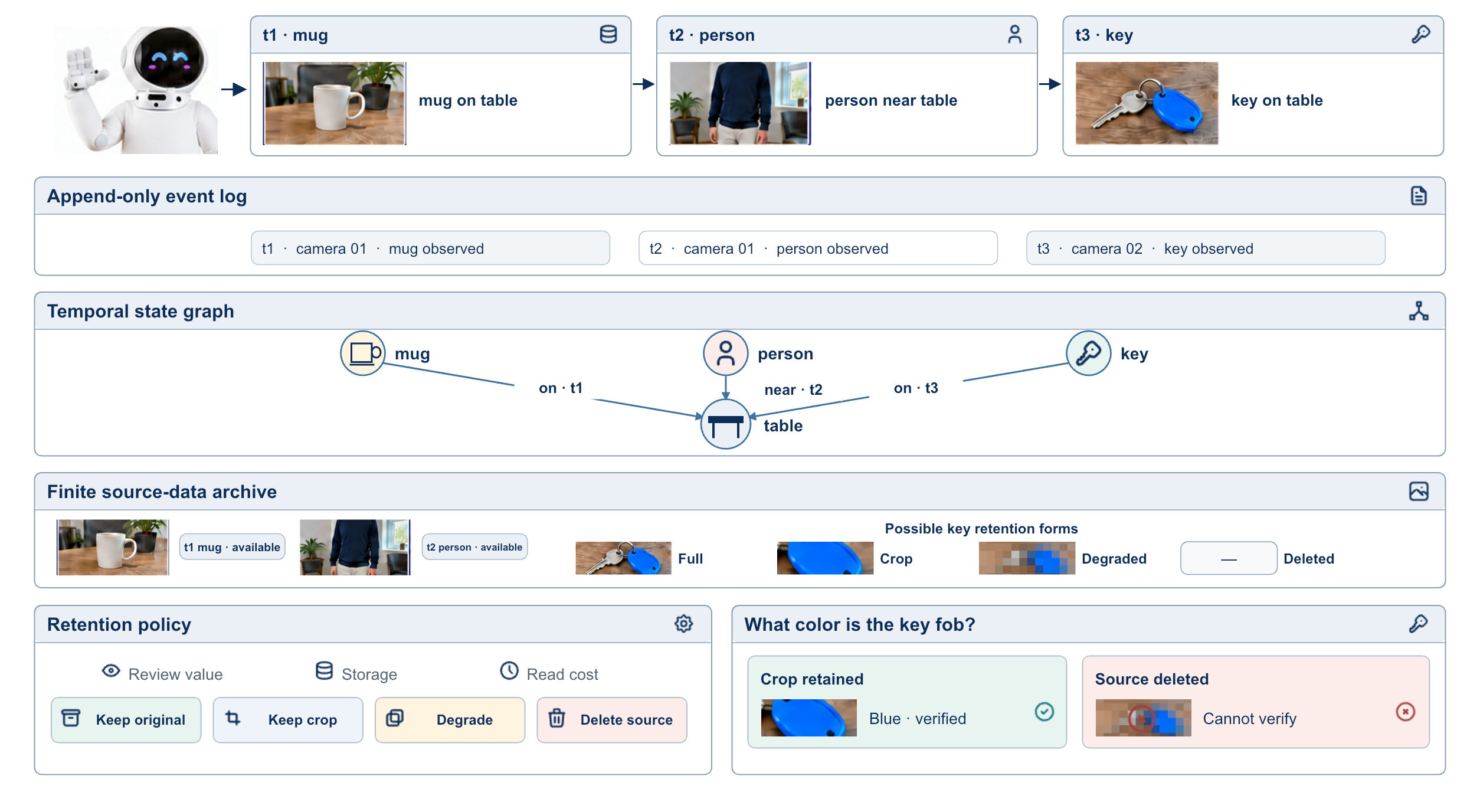}
\caption{Source retention and future answerability in an entryway scenario. An append-only event log and temporal state graph link compact records to a finite source archive. Keep, crop, degrade, and delete operations trade storage cost against future answerability. Retaining the key-fob crop preserves color evidence for later review; deleting the source removes that verification path.}
\label{fig:governance}
\end{figure}

\paragraph{Cost--utility allocation.}
\label{sec:governance}
Long-horizon memory allocates capacity between structured histories and inspectable source data, and allocates read resources between record retrieval and evidence review. These choices instantiate the separate budgets $B_s$, $B_e$, and $B_r$ in Section~\ref{sec:problem}. Event compression and source retention have different effects: compact summaries preserve semantic cues, while original images or audio preserve details for later verification. Event-centered compression has precedents in MemForest~\citep{wang2026memforest}, and a complete structured interaction log supports decision tracing as in PRO-LONG~\citep{fox2026prolong}. For item $i$, a candidate utility combines task relevance $R_i$, distinctiveness $D_i$, conflict-resolution value $C_i$, and estimated future review value $O_i$, less storage $S_i$ and access $A_i$ costs:
\begin{equation}
U_i=\alpha R_i+\beta D_i+\gamma C_i+\delta O_i-\lambda S_i-\mu A_i.
\end{equation}
The coefficients weight the utility and cost terms. The score ranks full-frame, crop, degraded, and summary-only retention choices under a byte limit, with future test queries excluded from retention decisions. Utility per byte provides a reference allocation rule; comparisons of stronger selectors require equal retained bytes and read cost. Atomic calls expose the resulting storage, inspection, update, and forgetting decisions together with their resource use. A recent buffer retains full observations temporarily. The finite archive then prioritizes state-changing interactions, uncertain identities, conflicting sources, and distinctive details, subject to capacity and permissions. Figure~\ref{fig:governance} illustrates the resulting answerability trade-off: a crop can preserve a key-fob color at lower cost than a full frame, while a summary or crop lacking that detail requires another source for verification. Retained frames are inspected only when required by the query and permitted by the read budget.

\paragraph{Source availability and consolidation.}
Each source exposes its retained, degraded, deleted, restricted, or temporarily offline status during retrieval, together with any permitted alternatives. Marking a state stale preserves the historical event, and merging duplicate summaries retains provenance. Permanent deletion requires authorization, invalidates dependent derivatives and indexes, and leaves a permitted minimal deletion record with its own lifetime. Importance- and time-sensitive forgetting provides a related retention principle~\citep{zhong2023memorybank}. Archive access latency counts toward the read budget. Consolidation merges duplicate descriptions, repairs indexes, groups episodes, and proposes regularities during low-load periods. It preserves temporal scope, uncertainty, and conflicting evidence, and uses the same versioned revision checks as online writing. The edge maintains recent observations, local identities, a query index, and interaction state. Authorized cloud consolidation receives bounded event fields and permitted evidence references, then returns candidates for edge-side validation against source links and read versions. Authenticated encryption, purpose limits, and retention limits govern cross-boundary access; cloud unavailability defers consolidation while local queries and revisions continue.

\section{Experiments}
\label{sec:experiments}

The experiments address four research questions connecting memory accuracy and reliability, recurring execution errors, and long-horizon resource costs:
\begin{itemize}
  \item \textbf{RQ1 (Recall and reasoning).} Can \betamem{} support accurate memory question answering, including multi-observation reasoning, unsupported questions, and habit recall, using a 9B model for memory generation? (Section~\ref{sec:exp-external})
  \item \textbf{RQ2 (Memory reliability).} How do write learning and source-linked components affect update usability and downstream answer accuracy? (Section~\ref{sec:training-effects})
  \item \textbf{RQ3 (Interactive execution).} Does memory-guided planning improve task success and reduce recurring errors in changing environments? (Section~\ref{sec:exp-emem})
  \item \textbf{RQ4 (Efficiency).} What latency, token, and storage costs does memory construction incur? (Section~\ref{sec:exp-efficiency})
\end{itemize}

\subsection{Experimental Setup}
\label{sec:exp-setup}
\label{sec:exp-data}
\label{sec:exp-metrics}

\paragraph{Datasets.}
We evaluate \betamem{} on four memory question-answering benchmarks and an interactive embodied task benchmark. M3-Bench-robot covers cross-modal reasoning (CM), general knowledge (GK), multi-detail reasoning (ME), multi-hop reasoning (MH), and person understanding (PU). We report judged accuracy on the full benchmark and by question type; All follows the benchmark's reported aggregation. LoCoMo contains ten long multi-session conversations and 1{,}986 questions across single-hop, temporal, multi-hop, open-domain, and adversarial categories. We report overall and category-level binary accuracy under the Mem0 judge-prompt protocol~\citep{chhikara2025mem0}. OpenEQA EM-EQA contains 1{,}636 questions over 152 ScanNet and HM3D indoor scenes, spanning seven capability types including object recognition, localization, spatial reasoning, functional reasoning, and world knowledge. Its LLM-Match metric converts five-level judge scores $s_i\in\{1,\dots,5\}$ to a normalized 0--100\% scale:
\[
M=\frac{1}{N}\sum_{i=1}^{N}\frac{s_i-1}{4}\times 100.
\]
We report LLM-Match for ScanNet, HM3D, and all questions combined. EgoLifeQA evaluates multi-day egocentric memory through entity logs, event recall, habit insight, relation maps, and task memory. We use its 500-question A1-JAKE subset and report four-choice exact-match accuracy overall and by type, without an LLM judge. EMem-Bench~\citep{liang2026embodiedmemorybench} contains 2{,}554 episodes: 1{,}036 Passive Observation, 1{,}052 Dynamic Tracking, 263 Interaction Failure, and 203 Experience Generalization episodes. These assess fine-grained visual memory, dynamic state tracking, interaction feedback, and experience generalization. Success rate (SR) measures whether actions reach the target state within the action budget. Error Recurrence Rate (ERR) uses the evaluator's action traces and frozen annotations to identify stale-location choices, repeated failed interactions, and violations of learned regularities. Avg SR and Avg ERR are equal-weight means over the four task families; ERR is distinct from write-service call errors. The executability of generated memory updates and the generation cost are evaluated on 20 test clips from our in-house dataset.

\paragraph{Baselines.}
Memory QA baselines cover full-context and direct-VQA models, streaming and structured visual memory, and agent-memory systems. Representative methods include M3-Agent and StreamMeCo on M3-Bench, Mem0 and ABot-AgentOS on LoCoMo, ConceptGraphs and 3D-Mem on OpenEQA, and EgoButler, WorldMM, and EGAgent on EgoLifeQA. Tables~\ref{tab:m3bench}--\ref{tab:egolife} give the complete comparisons and published sources. The memory-generation-model/method column lists only models used to generate or update memory in the cited experiment, or non-generative memory-construction methods. An em dash denotes an unreported memory-generation model or the absence of a separate memory-generation stage. For EMem-Bench, published vision-language and memory systems in Table~\ref{tab:emem} provide reference results. Memory generation evaluation compares a Qwen3.6-Flash teacher with base and SFT+GRPO-trained Qwen3.5-9B models on the same clips, keeping the surrounding service and QA configuration fixed.

\paragraph{Implementation Details.}
Across the four QA benchmarks, memory generation uses the native open-source Qwen3.5-9B model. Retrieval and answer generation operate on the constructed memories. Retrieval uses a hybrid method combining keyword matching and vector similarity search, while answer generation uses Qwen3.7-Plus. Historical memory construction does not access test questions. OpenEQA uses 24 frames per scene. Interactive execution uses Qwen3-VL-8B-Instruct. Historical observations and action logs are ingested before the task, and the planner receives the baseline context, a task-conditioned memory answer, current observations, and up to six retrieved keyframes. It generates action sequences within an eight-action simulator budget. EMem-Bench thus evaluates the combined retrieval-guidance and action-generation system. GRPO samples eight action lists per window at temperature 1.0, with a KL constraint relative to the SFT-initialized model and a maximum of 100 training steps. Implementation uses FSDP and a Hugging Face rollout backend in verl~\citep{sheng2024verl}. The teacher used for memory generation evaluation uses API version \texttt{qwen3.6-flash-2026-04-16}; base and trained models share the Qwen3.5-9B backbone.

\subsection{RQ1: Memory Question Answering}
\label{sec:exp-external}

We report \betamem{} on M3-Bench, LoCoMo,
OpenEQA EM-EQA, and EgoLifeQA in
Tables~\ref{tab:m3bench}--\ref{tab:egolife}.

\paragraph{M3-Bench.}
Table~\ref{tab:m3bench} reports \betamem{} scores on the benchmark's
five question types (Section~\ref{sec:exp-data}) and includes results reported by
\citet{long2025m3agent}, \citet{wang2026streammeco}, \citet{jiang2026nsmem}, and \citet{wang2026memforest}.

\begin{table}[!htbp]
\centering\scriptsize
\setlength{\tabcolsep}{3pt}
\caption{M3-Bench-robot results by question type (\%, higher is better).
All is the reported aggregate, not an unweighted average of the five
displayed categories.}
\label{tab:m3bench}
\begin{tabularx}{\linewidth}{@{}>{\raggedright\arraybackslash}p{0.34\linewidth}>{\raggedright\arraybackslash}Xrrrrrr@{}}
\toprule
Method & \mbox{Memory generation model/method} & All & CM & GK & ME & MH & PU \\
\midrule
Qwen2.5-Omni-7B \citep{long2025m3agent} & Qwen2.5-Omni-7B & 2.0 & 1.5 & 2.1 & 2.1 & 1.4 & 1.5 \\
Qwen2.5-VL-7B \citep{long2025m3agent} & Qwen2.5-VL-7B & 3.4 & 3.6 & 3.4 & 2.9 & 3.8 & 4.6 \\
Gemini-1.5-Pro \citep{long2025m3agent} & Gemini-1.5-Pro & 8.0 & 8.0 & 7.6 & 6.5 & 7.5 & 9.7 \\
GPT-4o \citep{long2025m3agent} & GPT-4o & 8.5 & 8.4 & 7.3 & 9.3 & 9.0 & 10.2 \\
MovieChat \citep{song2024moviechat} & Implicit feature memory & 11.2 & 12.2 & 7.0 & 13.3 & 9.8 & 15.7 \\
MA-LMM \citep{he2024malmm} & Implicit feature memory & 24.4 & 22.7 & 14.4 & 25.6 & 23.4 & 39.1 \\
Flash-VStream \citep{zhang2024flashvstream} & Implicit feature memory & 19.4 & 19.3 & 14.1 & 21.6 & 19.4 & 24.3 \\
Gemini-Agent \citep{long2025m3agent} & Gemini-1.5-Pro & 16.9 & 15.3 & 15.5 & 15.8 & 17.1 & 20.0 \\
Gemini-GPT4o-Hybrid \citep{long2025m3agent} & Gemini-1.5-Pro & 24.0 & 22.7 & 23.1 & 21.3 & 25.5 & 28.8 \\
M3-Agent \citep{long2025m3agent} & Qwen2.5-Omni-7B (SFT) & 30.7 & 31.2 & 19.1 & 32.8 & 29.4 & 43.3 \\
StreamForest \citep{zeng2025streamforest,wang2026streammeco} & Implicit feature memory & 32.1 & 31.3 & 19.0 & 33.4 & 29.4 & 44.3 \\
StreamMeCo + TMR (30\% compression) \citep{wang2026streammeco} & Qwen2.5-Omni-7B (SFT) & 34.6 & 35.5 & 27.5 & 35.7 & 32.9 & 44.0 \\
NS-Mem \citep{jiang2026nsmem} & SK-Gen / procedural DAG & 34.7 & 33.8 & 26.4 & 36.2 & 31.5 & 45.7 \\
MemForest + AGPR (30\%) \citep{wang2026memforest} & EventTree / GPT-4o-mini & 39.0 & 36.8 & \textbf{29.4} & 39.4 & 37.6 & 50.5 \\
\midrule
\betamem{} & Qwen3.5-9B & \textbf{48.08} & \textbf{47.52} & 22.08 & \textbf{55.36} & \textbf{61.11} & \textbf{78.43} \\
\bottomrule
\end{tabularx}
\end{table}

\betamem{} reaches 48.08 overall. On M3-Bench-robot, multi-detail, multi-hop, and person-understanding scores reach 55.36\%, 61.11\%, and 78.43\%, exceeding the strongest listed reference in each category by 15.96, 23.51, and 27.93 percentage points, respectively. All three margins exceed the 9.08-point overall margin. The larger gains on these categories point to the value of preserving relationships among observations: multi-detail questions require combining attributes, multi-hop questions require recovering event chains, and person-understanding questions require maintaining identity across encounters. The linked L0--L5 representation makes these records accessible through shared entity, temporal, and source references, allowing retrieval to assemble an evidence chain rather than return isolated similar fragments. This category profile connects the advantage to the central role of structured, reusable world memory.

\paragraph{LoCoMo.}
Table~\ref{tab:locomo} reports four non-adversarial question types and adversarial accuracy in one table. \betamem{} uses the Mem0-protocol judge prompt and covers 841 single-hop, 321 temporal, 282 multi-hop, and 96 open-domain questions. Its four-category aggregate is computed directly from 1{,}418 correct answers out of 1{,}540 questions.

\begin{table}[!htbp]
\centering\scriptsize
\setlength{\tabcolsep}{3pt}
\caption{LoCoMo LLM-Judge results by question type (\%). -- means unreported.}
\label{tab:locomo}
\begin{tabularx}{\linewidth}{@{}>{\raggedright\arraybackslash}p{0.34\linewidth}>{\raggedright\arraybackslash}Xrrrrrr@{}}
\toprule
Method & \mbox{Memory generation model/method} & SH & Temp & MH & Open & Adv & Overall \\
\midrule
Mem0 \citep{chhikara2025mem0} & GPT-4o-mini & 67.13 & 55.51 & 51.15 & 72.93 & -- & $66.88\!\pm\!0.15$ \\
Mem0 (Graph) \citep{chhikara2025mem0} & GPT-4o-mini & 65.71 & 58.13 & 47.19 & \textbf{75.71} & -- & $68.44\!\pm\!0.17$ \\
MIRIX \citep{wang2025mirix} & GPT-4.1-mini & 85.11 & 88.39 & 83.70 & 65.62 & -- & 85.38 \\
Zep \citep{hu2026evermemos,rasmussen2025zep} & GPT-4.1-mini & 90.84 & 77.26 & 81.91 & 75.00 & -- & 85.22 \\
EverMemOS \citep{hu2026evermemos} & GPT-4o-mini & 91.08 & 81.93 & 86.17 & 66.67 & -- & 86.76 \\
StructMem \citep{xu2026structmem} & DeepSeek-V3.2 & 85.14 & 77.57 & 73.05 & 53.12 & -- & 79.35 \\
ABot-AgentOS Static \citep{tian2026abotagentos} & Qwen3.6-Plus & 92.9 & 87.5 & 90.9 & 70.8 & 80.9 & 87.5 \\
\midrule
\betamem{} & Qwen3.5-9B & \textbf{95.2} & \textbf{90.3} & \textbf{95.4} & 60.4 & \textbf{89.9} & \textbf{92.08} \\
\bottomrule
\end{tabularx}
\end{table}

LoCoMo adversarial questions test recognition of requests that cannot be answered from the dialogue~\citep{maharana2024locomo}. \betamem{} scores 89.9\% on the 446 adversarial questions; its 92.08\% overall in the table covers the 1{,}540 non-adversarial questions. The combination of strong non-adversarial accuracy and adversarial rejection addresses two complementary requirements: recovering recorded information and recognizing when the available history does not support an answer. Linking retrieved records to their sources gives the answerer a basis for checking a question's premise instead of treating semantic relevance as evidence of truth. This is the conversational counterpart of the distinction between a historical search lead and a verified current state in embodied memory.

\paragraph{OpenEQA EM-EQA.}
Table~\ref{tab:openeqa} reports LLM-Match by scene source for methods with a separate memory-construction stage. \betamem{} uses a 24-frame budget for 1{,}079 ScanNet and 557 HM3D questions.

\begin{table}[!htbp]
\centering\scriptsize
\setlength{\tabcolsep}{4pt}
\caption{OpenEQA EM-EQA LLM-Match of memory methods by scene source (higher is better).
-- means unreported. Frame budgets differ across methods.}
\label{tab:openeqa}
\begin{tabularx}{\linewidth}{@{}>{\raggedright\arraybackslash}p{0.34\linewidth}>{\raggedright\arraybackslash}Xrrrr@{}}
\toprule
 & & & \multicolumn{3}{c}{LLM-Match (\%)} \\
\cmidrule(lr){4-6}
Method & \mbox{Memory generation model/method} & Frames & ScanNet & HM3D & Overall \\
\midrule
ConceptGraphs \citep{gu2023conceptgraphs,tian2026abotagentos} & LLaVA-7B & 10 & 37.8 & 34.0 & 36.5 \\
R-EQA \citep{ong2025reqa,tian2026abotagentos} & Qwen2.5-VL-7B & 3 & 49.1 & 42.8 & 46.0 \\
GraphPad \citep{ali2025graphpad,tian2026abotagentos} & Gemini-2.0-Flash & 5--20 & -- & -- & 55.3 \\
3D-Mem / SnapMem \citep{yang2025threedmem,tian2026abotagentos} & Visual snapshots & 3.1 & -- & -- & 57.2 \\
GaussExplorer \citep{kim2026gaussexplorer,tian2026abotagentos} & 3D reconstruction & -- & -- & -- & 57.8 \\
ABot-AgentOS Static \citep{tian2026abotagentos} & GPT-5.4 & 24 & 61.9 & 55.7 & 59.9 \\
ABot-AgentOS + Self-evo \citep{tian2026abotagentos} & GPT-5.4 & 24 & -- & -- & 60.4 \\
\midrule
\betamem{} & Qwen3.5-9B & 24 & \textbf{73.7} & \textbf{70.4} & \textbf{72.6} \\
\bottomrule
\end{tabularx}
\end{table}

\betamem{} scores 73.7\% on ScanNet and 70.4\% on HM3D, yielding
72.6\% overall on the LLM-Match scale. With the same 24-frame count, the GPT-5.4 memory reference ABot-AgentOS Static scores 59.9\%. Indoor-scene questions depend on details such as object attributes, spatial relations, and open/closed states that compact semantic records may omit. The separation between structured memory retrieval and source-evidence access lets records locate the relevant observation while retained visual evidence supplies the detail needed for the answer. The results therefore connect scene understanding to the joint use of semantic organization and inspectable sources under a finite visual-read budget.

\paragraph{EgoLifeQA.}
Table~\ref{tab:egolife} reports four-choice accuracy on the A1-JAKE
subset (Section~\ref{sec:exp-data}) by question type. The \betamem{}
breakdown covers 125 EntityLog, 126 EventRecall, 61 HabitInsight,
125 RelationMap, and 63 TaskMaster questions.

\begin{table}[!htbp]
\centering\scriptsize
\setlength{\tabcolsep}{3pt}
\caption{EgoLifeQA four-choice accuracy by question type (\%, higher
is better). EL, ER, HI, RM, and TM denote EntityLog, EventRecall,
HabitInsight, RelationMap, and TaskMaster. \betamem{} uses A1-JAKE
(500 questions).}
\label{tab:egolife}
\begin{tabularx}{\linewidth}{@{}>{\raggedright\arraybackslash}p{0.34\linewidth}>{\raggedright\arraybackslash}Xrrrrrr@{}}
\toprule
Method & \mbox{Memory generation model/method} & EL & ER & HI & RM & TM & Overall \\
\midrule
Qwen3.5-Flash, direct VQA (64 frames) \citep{tian2026abotagentos} & Qwen3.5-Flash & 42.4 & 39.7 & 45.9 & 39.2 & 46.0 & 41.8 \\
Gemini-2.5-Pro, direct VQA (3,000 frames) \citep{rege2026egagent} & Gemini-2.5-Pro & 45.6 & 48.4 & 51.7 & 41.6 & 52.4 & 46.8 \\
EgoGraph \citep{sun2026egograph} & InternVL-3.5-8B / GPT-4o & 46.4 & 46.8 & 45.9 & 35.2 & 60.3 & 45.8 \\
EgoButler + Gemini-1.5-Pro captioner \citep{yang2025egolife} & Gemini-1.5-Pro & 36.0 & 37.3 & 45.9 & 30.4 & 34.9 & 36.9 \\
EgoButler + GPT-4o captioner \citep{yang2025egolife} & GPT-4o & 34.4 & 42.1 & 29.5 & 30.4 & 44.4 & 36.2 \\
EgoButler-EgoGPT \citep{yang2025egolife,tian2026abotagentos} & EgoGPT-7B & 39.2 & 36.5 & 31.1 & 33.6 & 39.7 & 36.0 \\
WorldMM \citep{yeo2025worldmm,tian2026abotagentos} & Qwen3.5-Flash & 49.6 & 54.8 & 54.1 & \textbf{62.4} & 60.3 & 56.0 \\
EGAgent \citep{rege2026egagent,tian2026abotagentos} & Gemini 2.5 Pro  & 54.4 & 57.1 & 60.3 & \textbf{62.4} & 74.6 & 57.5 \\
\midrule
\betamem{} & Qwen3.5-9B & \textbf{55.2} & \textbf{62.7} & \textbf{65.6} & 56.8 & \textbf{84.1} & \textbf{62.4} \\
\bottomrule
\end{tabularx}
\end{table}

\betamem{} achieves 62.4\% overall exact-match accuracy. On EgoLifeQA A1-JAKE, HabitInsight accuracy reaches 65.6\%, exceeding the strongest listed reference in that category by 5.3 percentage points. Habit questions require recovering regularities across episodes, making reusable summaries of repeated experience relevant to individual routines. The result is consistent with connecting episodic observations to person-linked habit records. Recovering such regularities provides a basis for personalized memory that retains an individual's routines and preferences. EventRecall and TaskMaster scores of 62.7\% and 84.1\% further connect this ability to recovering event histories and task-relevant experience. Together, these categories motivate linking episodic records with reusable experiential and procedural information: a useful memory must retain both what happened in a particular episode and what repeated episodes reveal about a person. RelationMap remains below the strongest listed references, identifying relational organization as a distinct area for improvement.

\paragraph{Overall performance.}
\label{sec:exp-insights}
Across Tables~\ref{tab:m3bench}--\ref{tab:egolife}, \betamem{} using Qwen3.5-9B for memory generation achieves the highest overall scores among the listed memory methods under the stated evaluation protocols. The shared architectural contribution is the organization of observations into temporally scoped, source-linked records that can be reused across these task settings. The category-level analyses above connect this organization to compositional reasoning, evidence-supported answering, scene-detail recovery, and person-linked regularities. Qwen3.5-9B performs memory generation, while retrieval and answer generation operate on the resulting records; the results characterize the complete memory pipeline rather than the isolated reasoning capacity of the 9B memory generation model.

\begin{figure*}[!htbp]
\centering
\includegraphics[width=\textwidth]{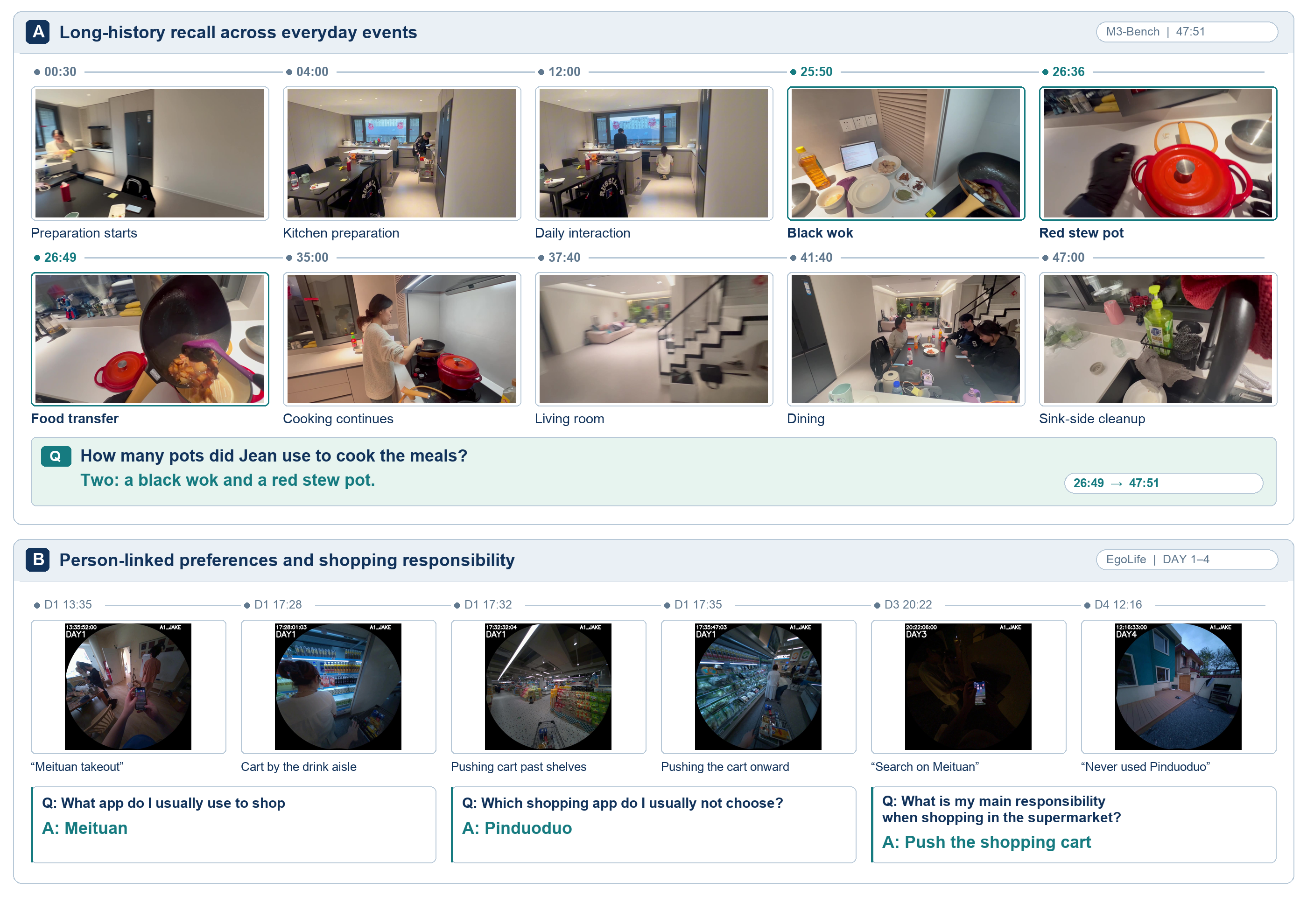}
\caption{Long-history and person-linked memory examples. (A) Recalling cookware after approximately 48 minutes of activity history. (B) Answering questions about platform choices and shopping responsibility from experiences across days. All evidence shown precedes the corresponding query time.}
\label{fig:qualitative-cases}
\end{figure*}

\paragraph{Qualitative examples.}
Figure~\ref{fig:qualitative-cases} presents two examples of recalling earlier events and personal experiences. In M3-Bench, a question is asked at 47:51. By retrieving a pot-changing sequence from about 21 minutes earlier, \betamem{} correctly answers that two pots were used for cooking. In EgoLifeQA, three independent questions ask which shopping platform Jake usually uses, which platform he does not use, and his main responsibility when shopping in the supermarket; the figure presents the corresponding reference answers from the original dataset. These examples show how retaining event details and connecting experiences across days can support question answering.

\subsection{RQ2: Memory Reliability and Evidence Grounding}
\label{sec:training-effects}
We examine write learning through update usability and training dynamics, then use component ablations to test how source-linked memory affects subsequent answers.

\begin{figure}[!htbp]
\centering
\includegraphics[width=\linewidth]{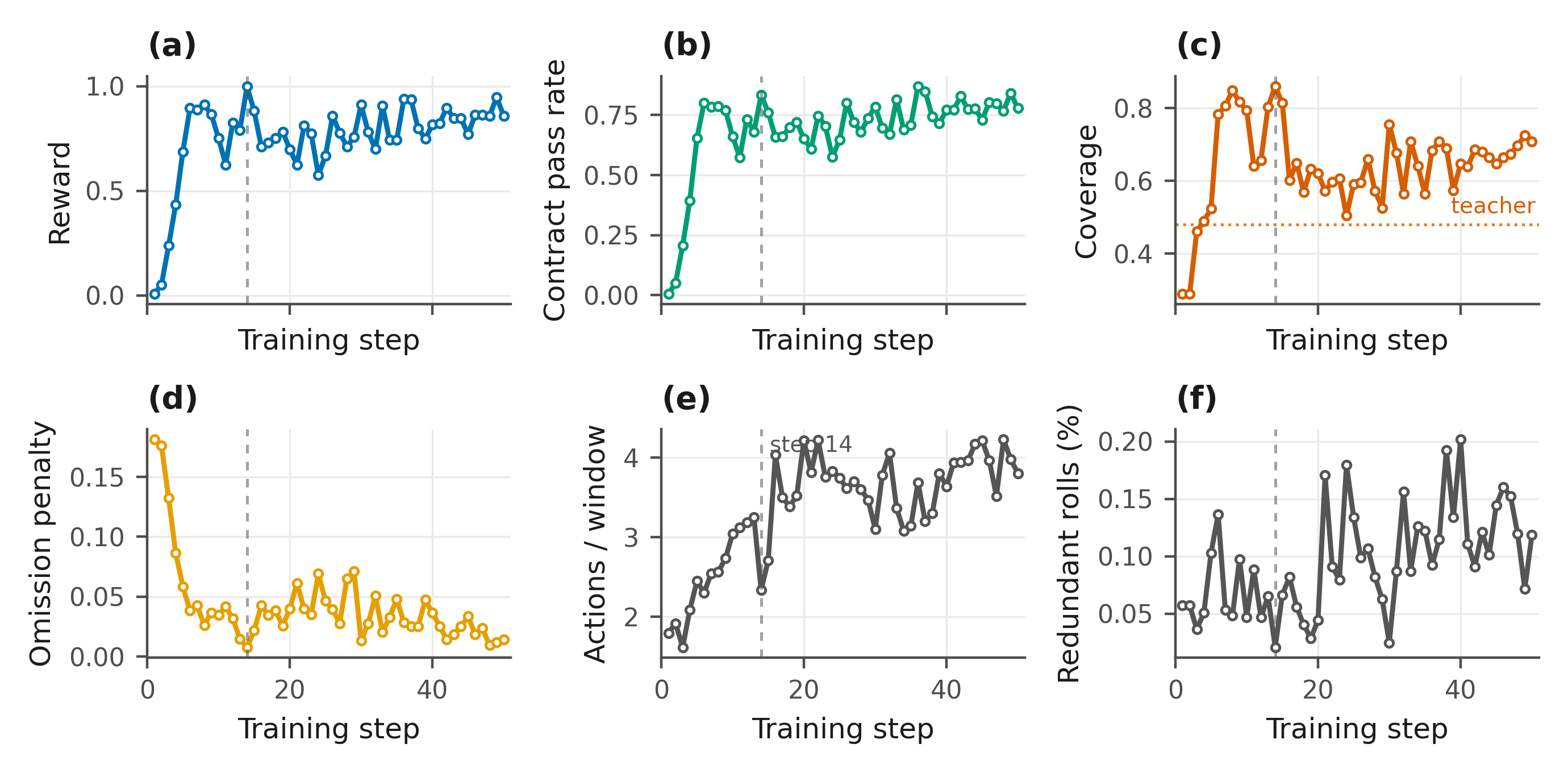}
\caption{Training dynamics of the write policy over 50 logged steps, from per-roll reward-component logs of the reward implementation. (a)~Mean reward; (b)~contract pass rate; (c)~coverage of the should-record sets, with the dotted line marking the teacher-pool reference (0.48); (d)~omission penalty; (e)~actions per window; (f)~share of rolls containing redundant location updates. The vertical dashed line marks the step-14 checkpoint, where coverage is relatively high and redundant updates are relatively infrequent. Coverage exceeds the teacher-pool reference within four steps.}
\label{fig:training-dynamics}
\end{figure}

\paragraph{Training dynamics.}
\label{sec:training-optimization}
GRPO training uses the teacher-replay-verified coverage targets and reward formulation described in Section~\ref{sec:training}. The training curves and operational results were obtained using the reward implementation described in Section~\ref{sec:training}. We log parseability, coverage, complete omissions, duplicate and redundant actions, action count, and total reward. Target coverage rises from 0.29 to 0.86 and passes the teacher dataset's 0.48 reference level within four steps; the omission penalty falls from 0.18 to 0.008 (Figure~\ref{fig:training-dynamics}). Component-level logging distinguishes improved coverage from reward growth accompanied by redundant updates.

\paragraph{Write usability and generation cost.}
Contract pass rate measures compliance with the memory store's rules. A window is \emph{emptied by repair} when no proposed action remains after checking and repair; service-call errors are measured separately. Output tokens per stored action counts all generated output, including discarded actions, relative to the actions that reach the store.

\begin{table}[!htbp]
\centering\small
\caption{Memory-writing behavior on 20 test video clips from our
in-house dataset. All models use these clips.}
\label{tab:write-ops}
\setlength{\tabcolsep}{4pt}
\begin{tabularx}{\linewidth}{@{}Yrrr@{}}
\toprule
 & Teacher & Native & SFT+GRPO \\
\midrule
Base model & Qwen3.6-Flash & Qwen3.5-9B & Qwen3.5-9B \\
\midrule
\multicolumn{4}{l}{\emph{Action usability}} \\
Contract pass rate (\%) & 72.0 & 50.9 & 86.2 \\
Windows emptied by repair (\%) & 6.2 & 20.8 & 3.1 \\
Service call error rate (\%) & 0.18 & 1.58 & 0.18 \\
\midrule
\multicolumn{4}{l}{\emph{Generation cost}} \\
Output tokens per stored action & 149 & 224 & 118 \\
\bottomrule
\end{tabularx}
\end{table}

Training raises contract pass rate from 50.9\% to 86.2\% (+35.3 percentage points), above the teacher's 72.0\%, while reducing repair-emptied windows from 20.8\% to 3.1\% and service-call errors from 1.58\% to 0.18\%. Observed failures of the base model include timestamps expressed as strings. Output tokens per stored action fall from 224 to 118 (47.3\% lower), compared with 149 for the teacher. The trained model thus produces accepted memory updates with less discarded output and lower generation overhead. Higher update acceptance and fewer repair-emptied windows mean that less generated content is lost before entering memory. The simultaneous reduction in output tokens per stored action links interface-level learning to ingestion efficiency: the model learns to express useful updates through executable atomic operations with less corrective processing. This directly supports the write-contract contribution, alongside the source-linked component analysis below.

\paragraph{Evidence-linked component ablations.}
\label{sec:exp-ablation}
We evaluate three memory components on M3-Bench-robot. Memory generation uses Qwen3.5-9B. Retrieval uses a hybrid method combining keyword matching and vector similarity search, while answer generation uses Qwen3.7-Plus. We report the Full configuration and the QA accuracy and drop when each component is removed separately.

\begin{table}[!htbp]
\centering\small
\setlength{\tabcolsep}{5pt}
\caption{Ablation of three memory components. Each ablated configuration removes one component from Full. QA is accuracy; drops are percentage-point differences from the displayed Full score.}
\label{tab:memory-ablation}
\begin{tabularx}{\linewidth}{@{}Xrr@{}}
\toprule
Configuration & QA (\%)$\uparrow$ & Drop (pp) \\
\midrule
\textbf{Full architecture (Full)} & \textbf{48.08} & -- \\
w/o linked source utterances & 38.61 & 9.47 \\
w/o visual entity discovery and confirmation & 42.69 & 5.39 \\
w/o visual--event consistency verification & 42.31 & 5.77 \\
\bottomrule
\end{tabularx}
\end{table}

\paragraph{Semantic speech compression with linked source evidence.}
Speech is compressed into retrievable semantic summaries linked to timestamped, speaker-attributed utterance evidence. Summaries provide compact semantic cues, while source utterances support on-demand checks of wording, speakers, and temporal details. Compared with retaining semantic summaries alone, restoring linked utterance evidence raises QA from 38.61\% to 48.08\%, a gain of 9.47 percentage points. This supports the complementary use of semantic compression and source evidence.

\paragraph{Visual entity discovery and confirmation.}
The main-window analysis proposes candidates, an independent scan recovers omissions, and local category correction and confirmation rules establish entity records. Without this mode, the main analysis still generates candidates, but lacks the supplementary scan and applies more conservative confirmation. Restoring the combined discovery and confirmation component raises QA from 42.69\% to 48.08\%, a gain of 5.39 percentage points.

\paragraph{Visual--event consistency verification.}
Independent checks verify candidate categories against visual evidence and resolve conflicts between entity records and event descriptions, reducing erroneous memory writes. Restoring this component raises QA from 42.31\% to 48.08\%, a gain of 5.77 percentage points, supporting visual--event consistency checks during memory writing.

The three removals target complementary failure paths: losing utterance detail, missing or misidentifying entities, and admitting visually inconsistent events. Their respective accuracy drops show that memory quality depends on evidence preservation as well as validation before updates. The benefit therefore extends beyond producing well-formed tool calls to constructing records that remain useful for downstream reasoning.

\subsection{RQ3: Interactive Execution and Error Recurrence}
\label{sec:exp-emem}
We evaluate memory-guided planning on tasks involving changing states, failed interactions, and experience generalization. The error-recurrence metric quantifies failure patterns that motivate policy adaptation in Section~\ref{sec:evolution}.
Table~\ref{tab:emem} reports the Qwen3-VL-8B-Instruct evaluation and published reference results. On EMem-Bench, memory-guided planning achieves an average success rate of 38.4\% and an average error recurrence rate of 24.9\%. Dynamic Tracking SR is 60.7\% and ERR is 23.9\%; Experience Generalization SR is 35.0\%. These results characterize memory-supported action selection across the task families. \betamem{} is particularly effective at tracking state in changing scenes and using historical observations to support subsequent action selection.

\begin{table}[!htbp]
\centering\scriptsize
\setlength{\tabcolsep}{2pt}
\caption{EMem-Bench results (\%). Published results are from \citet{liang2026embodiedmemorybench}. Each group separately reports success rate (SR$\uparrow$) and error recurrence rate (ERR$\downarrow$); Average is the equal-weight mean over the four task families. Dataset composition is given in Section~\ref{sec:exp-data}. Qwen3-VL models are Instruct variants.}
\label{tab:emem}
\begin{tabularx}{\linewidth}{@{}>{\raggedright\arraybackslash}X>{\raggedright\arraybackslash}p{0.17\linewidth}rrrrrrrrrr@{}}
\toprule
System & Base model & \multicolumn{2}{c}{Average} & \multicolumn{2}{c}{Passive} & \multicolumn{2}{c}{Dynamic} & \multicolumn{2}{c}{Interaction} & \multicolumn{2}{c}{Exp. Gen.} \\
\cmidrule(lr){3-4}\cmidrule(lr){5-6}\cmidrule(lr){7-8}\cmidrule(lr){9-10}\cmidrule(lr){11-12}
 & & SR$\uparrow$ & ERR$\downarrow$ & SR$\uparrow$ & ERR$\downarrow$ & SR$\uparrow$ & ERR$\downarrow$ & SR$\uparrow$ & ERR$\downarrow$ & SR$\uparrow$ & ERR$\downarrow$ \\
\midrule
Qwen3-VL-4B-Ins \citep{liang2026embodiedmemorybench} & Qwen3-VL-4B & 30.1 & 37.4 & 46.4 & 36.0 & 37.3 & 45.4 & 20.9 & 22.8 & 15.8 & 45.3 \\
Qwen3-VL-8B-Ins \citep{liang2026embodiedmemorybench} & Qwen3-VL-8B & 25.4 & 41.1 & 41.4 & 45.0 & 26.6 & 58.9 & 11.8 & 22.1 & 21.7 & 38.4 \\
InternVL3-8B \citep{liang2026embodiedmemorybench} & InternVL3-8B & 15.5 & 48.8 & 20.5 & 62.4 & 31.8 & 55.9 & 5.3 & \textbf{10.3} & 4.4 & 66.5 \\
InternVL3-38B \citep{liang2026embodiedmemorybench} & InternVL3-38B & 31.8 & 47.4 & 47.9 & 33.7 & 17.0 & 64.5 & 29.3 & 29.3 & 33.0 & 62.1 \\
Ovis2-16B \citep{liang2026embodiedmemorybench} & Ovis2-16B & 29.0 & 48.0 & 44.5 & 43.8 & 22.7 & 74.2 & \textbf{30.0} & 36.1 & 18.7 & 37.9 \\
RynnBrain-8B \citep{liang2026embodiedmemorybench} & RynnBrain-8B & 3.6 & 37.5 & 5.3 & 38.9 & 4.0 & 46.6 & 2.7 & 28.1 & 2.5 & 36.5 \\
Cambrian-S-7B \citep{liang2026embodiedmemorybench} & Cambrian-S-7B & 0.8 & 61.4 & 0.8 & 83.3 & 0.9 & 67.2 & 1.1 & 36.5 & 0.5 & 58.6 \\
RoboBrain2.5-8B-NV \citep{liang2026embodiedmemorybench} & RoboBrain2.5-8B-NV & 20.0 & 42.3 & 34.1 & 57.4 & 29.4 & 68.4 & 16.4 & 25.1 & 0.0 & \textbf{18.2} \\
EMem \citep{liang2026embodiedmemorybench} & Qwen3-VL-8B & 36.2 & 28.3 & 44.1 & 39.0 & 57.1 & 26.2 & 16.0 & 17.5 & 27.6 & 30.5 \\
MIRIX \citep{liang2026embodiedmemorybench} & GPT-5.4-mini & 28.2 & 46.7 & 49.2 & 39.7 & 20.0 & 77.9 & 11.0 & 20.5 & 32.5 & 48.8 \\
MemVerse \citep{liang2026embodiedmemorybench} & GPT-5.4-mini & 28.9 & 45.1 & \textbf{49.6} & 39.6 & 18.1 & 70.1 & 14.5 & 24.0 & 33.5 & 46.8 \\
MMA \citep{liang2026embodiedmemorybench} & GPT-5.4-mini & 23.4 & 47.2 & 43.8 & 38.7 & 15.5 & 76.1 & 10.3 & 22.8 & 24.1 & 51.2 \\
\midrule
\betamem{} & Qwen3-VL-8B & \textbf{38.4} & \textbf{24.9} & 48.8 & \textbf{24.2} & \textbf{60.7} & \textbf{23.9} & 9.1 & 22.4 & 35.0 & 29.0 \\
\bottomrule
\end{tabularx}
\end{table}

\subsection{RQ4: Resource Costs over Long Horizons}
\label{sec:exp-efficiency}
We profile the stage-level costs underlying the memory budget allocation problem in Section~\ref{sec:governance}. The efficiency evaluation is based on logs of the native Qwen3.5-9B service on M3-Bench-robot, covering 100 videos and 1{,}277 questions. Grouping the sampled frames from the same videos into 10-frame windows yields an estimated 30{,}001 windows (about 300 per video); the table reports latency estimates for 10-frame windows. Window analysis extracts observations; entity scanning identifies
objects and people; memory updates ground, verify, and reconcile
records; experience summarization runs only when triggered. Stage latency and tokens are projected per call.
Videos average about 68 object entities, 580 events, and four experience summaries. Memory content occupies approximately 0.0571\,GB per 30 minutes on average.

\begin{table}[!htbp]
\centering\small
\caption{Core-stage efficiency calculated from the native Qwen3.5-9B
baseline on M3-Bench-robot for 10-frame windows. Stage mean, p95, and tokens (input/output) are calculated
per call; ingestion totals sum stage latency and tokens weighted by calls per window. The total in the p95 column is the call-weighted sum of stage p95 values. The token column reports mean consumption.}
\label{tab:efficiency}
\begin{tabular}{lcccc}
\toprule
Core stage & Calls/window & Mean & p95 & Mean tokens (in/out) \\
\midrule
Window analysis & 1.03 & 4.33\,s & 6.84\,s & 5.2k / 0.54k \\
Entity scanning & 0.99 & 1.83\,s & 4.52\,s & 4.5k / 0.20k \\
Memory updates & 1.75 & 0.63\,s & 1.23\,s & 1.0k / 0.042k \\
Experience summarization & 0.13 & 2.08\,s & 4.00\,s & 7.0k / 0.20k \\
\midrule
Ingestion total (per window) & 3.9 & 7.64\,s & 14.19\,s & 12.47k / 0.854k \\
\bottomrule
\end{tabular}
\end{table}

Window analysis has the highest projected mean latency per call, while experience summarization uses the most input tokens per call but runs less frequently. These measurements show why resource allocation must account for both call frequency and per-call cost. The infrequently triggered summary stage illustrates why a costly individual operation can still have a limited aggregate cost. Atomic interfaces expose both invocation frequency and per-call expenditure, enabling budgets to be allocated to operations that contribute useful evidence. This operational view grounds the cost--utility formulation in measurable storage and read decisions, alongside the reduction in generation cost achieved by write learning.

\section{Conclusion}
\label{sec:conclusion}
\label{sec:discussion}
\betamem{} brings together time-scoped, source-linked memory, evidence-constrained memory generation, and budgeted memory operations, while formulating failure-driven policy RSI through causal diagnosis, local rule repair, and independent replay. Its central insight is that useful long-horizon memory depends on maintaining the relationships among what was observed, when it was valid, and how it can be checked and revised as experience accumulates. The benchmark and component analyses support the value of this organization for connecting observations, recovering event histories, and recalling person-specific regularities, while training the memory generation model improves the reliability and efficiency of memory construction. These findings motivate treating memory as an evolving evidence-maintenance process within embodied agents. Looking ahead, this perspective suggests a path toward continual learning and personalization in changing environments, where agents adapt both their stored knowledge and the policies that manage it, subject to verifiable evidence and finite resources.

\paragraph{Privacy and Safety Issues.}
Person-linked memory raises privacy concerns. Embodied agents should verify safety-critical information before using memory to guide physical actions. Deployment tests should cover deletion, access control, and state validity alongside QA quality.

\paragraph{Limitations.}
Entity association, map alignment, and visual interpretation can fail together. Graph scoping can exclude the right evidence; original observations may still lack the desired detail. A semantic validator can agree with an erroneous proposer. Predicted review value may be unreliable under distribution shift, and finite storage can remove details needed by future queries. Answers depend on retained observations and structured records; current-state confirmation requires new observations.

\FloatBarrier
\bibliographystyle{plainnat}
{\footnotesize
\setlength{\bibsep}{0pt}
\bibliography{references}
}
\end{document}